\documentclass[10pt,journal,compsoc]{IEEEtran}
\ifCLASSOPTIONcompsoc
  \usepackage[nocompress]{cite}
\else
  \usepackage{cite}
\fi
\ifCLASSINFOpdf
  \usepackage[pdftex]{graphicx}
\else
  \usepackage[dvips]{graphicx}
\fi
\usepackage{amsmath}
\usepackage{amsthm}
\usepackage{amssymb}
\usepackage{enumitem}
\newtheorem{theorem}{Theorem}
\newtheorem{prop}{Proposition}

\usepackage{multirow}

\newcommand{\ie}{\emph{i.e., }}
\newcommand{\eg}{\emph{e.g., }}

\ifCLASSOPTIONcompsoc
\usepackage[caption=false,font=normalsize,labelfon
t=sf,textfont=sf]{subfig}
\else
\usepackage[caption=false,font=footnotesize]{subfi
g}
\fi

\begin{document}
%
\title{FFHR: Fully and Flexible Hyperbolic Representation for Knowledge Graph Completion}
%
%
%
%

\author{Wentao Shi, Junkang Wu, Xuezhi Cao, Jiawei Chen, Wenqiang Lei, Wei Wu, Xiangnan He,~\IEEEmembership{Senior Member, IEEE} 
\IEEEcompsocitemizethanks{\IEEEcompsocthanksitem W. Shi, J. Wu, J. Chen and X. He was with the School of Information Science and Technology, University of Science and Technology of China.\protect\\
Email:\{shiwentao123, jkwu0909\}@mail.ustc.edu.cn, cjwustc@ustc.edu.cn, xiangnanhe@gmail.com.
\IEEEcompsocthanksitem X. Cao and W. Wu were with meituan. \protect \\
Email:caoxuezhi@meituan.com, wuwei19850318@gmail.com.
\IEEEcompsocthanksitem W. Lei was with Sichuan University. \protect \\
Email:wenqianglei@gmail.com}
\thanks{Manuscript received April 19, 2005; revised August 26, 2015.}}

%
%

\markboth{Journal of \LaTeX\ Class Files,~Vol.~14, No.~8, August~2015}%
{Shell \MakeLowercase{\textit{et al.}}: Bare Demo of IEEEtran.cls for Computer Society Journals}
%



\IEEEtitleabstractindextext{%
\begin{abstract}
    Learning hyperbolic embeddings for knowledge graph (KG) has gained increasing attention due to its superiority in capturing hierarchies. However, some important operations in hyperbolic space still lack good definitions, making existing methods unable to fully leverage the merits of hyperbolic space. Specifically, they suffer from two main limitations: 1) existing Graph Convolutional Network (GCN) methods in hyperbolic space rely on tangent space approximation, which would incur approximation error in representation learning, and 2) due to the lack of inner product operation definition in hyperbolic space, existing methods can only measure the plausibility of facts (links) with hyperbolic distance, which is difficult to capture complex data patterns.
    
    In this work, we contribute: 1) a Full Poincar\'{e} Multi-relational GCN that achieves graph information propagation in hyperbolic space without requiring any approximation, and 2) a hyperbolic generalization of Euclidean inner product that is beneficial to capture both hierarchical and complex patterns. On this basis, we further develop a \textbf{F}ully and \textbf{F}lexible \textbf{H}yperbolic \textbf{R}epresentation framework (\textbf{FFHR}) that is able to transfer recent Euclidean-based advances to hyperbolic space. We demonstrate it by instantiating FFHR with four representative KGC methods. Extensive experiments on benchmark datasets validate the superiority of our FFHRs over their Euclidean counterparts as well as state-of-the-art hyperbolic embedding methods.
\end{abstract}

}

\maketitle

\IEEEdisplaynontitleabstractindextext

%
\IEEEpeerreviewmaketitle

\IEEEraisesectionheading{\section{Introduction}\label{sec:introduction}}

%
%
%
%
\IEEEPARstart{K}{nowledge} graphs have benefited lots of downstream tasks, such as dialogue generation \cite{He2017LearningSC}\cite{tuan2019dykgchat},  recommendation systems \cite{10.1145/3292500.3330989}\cite{9216015} and question answering \cite{10.1145/3289600.3290956}. Although KGs already contain tremendous entities and facts, they are still far from complete. This shortcoming has raised a hot research topic ---  \textit{Knowledge Graph Completion}, which aims to discover the missing facts in a given KG.

\begin{figure}[!t]
\centering
\includegraphics[width=0.95\columnwidth]{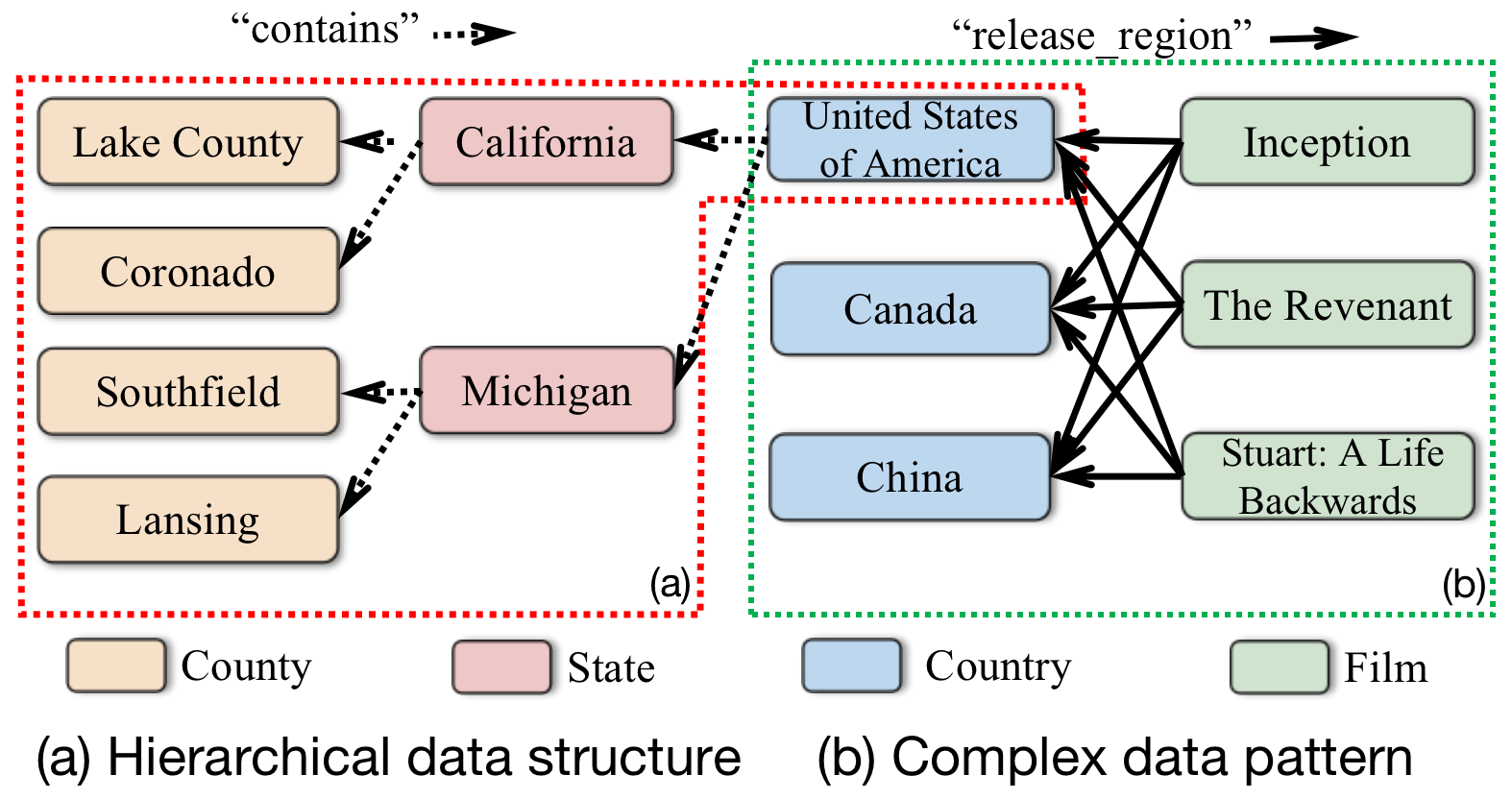}
\caption{A toy example of knowledge graph that simultaneously contains hierarchical data structure and complex data patterns.}
\label{figure_toy}
\end{figure}

Embedding methods are prevalent for KGC, which map entities and relations into a latent space and use a score function to measure the plausibility of facts. Recent research efforts mainly focused on embedding KGs in Euclidean space. However, as KGs are usually organized in a hierarchical structure (Fig. \ref{figure_toy} gives an example), the Euclidean space that makes no assumptions about the inherent geometric pattern of data would fall short in preserving such hierarchies. Depicting hierarchies in Euclidean space requires extremely high dimensional embeddings, which inevitably incurs high time and memory costs \cite{sun2018rotate}\cite{pmlr-v80-lacroix18a}.

More recently, hyperbolic space has been leveraged for KGC \cite{balazevic2019multi}\cite{chami2020low}, offering exciting opportunities to learn low-dimensional embeddings while preserving latent hierarchies. Hyperbolic space, which can be considered as a continuous analog of a discrete tree, has a powerful capacity in capturing hierarchical patterns--- e.g., it can embed arbitrary trees with arbitrarily low distortion in just two dimensions \cite{pmlr-v80-sala18a}. Despite promising results, we argue that existing methods have not fully exploited hyperbolic space. Many important operations in hyperbolic space still lack good definitions, making models fail to inherit the merits of recent advanced Euclidean-based models. Concretely, they suffer from two main limitations:

(1)  The effectiveness of GCN has been validated for KGC \cite{DBLP:conf/esws/SchlichtkrullKB18}\cite{vashishth2020compositionbased}. In order to convert GCN into hyperbolic space, existing methods \cite{10.1145/3442381.3450118}\cite{2202.13852} often rely on tangent space for feature transformation and neighborhood aggregation. However, as tangent space is just a local approximation of hyperbolic space, performing GCN via tangent space would inevitably incur approximation error, which would further distort the global structure of hyperbolic manifold.

(2) Existing hyperbolic methods always use hyperbolic distance as the score function, which is usually insufficient to capture complex relations (Fig. \ref{figure_toy}b). In fact, for the methods in Euclidean space, state-of-the-arts are often achieved by the methods with the inner product score function. Inner product is more flexible and effective, but it still lacks a clear corresponding definition in hyperbolic space.

\begin{figure}[t]
\centering
\subfloat[Classical Hyperbolic GCNs]{
        \includegraphics[width=0.9\columnwidth]{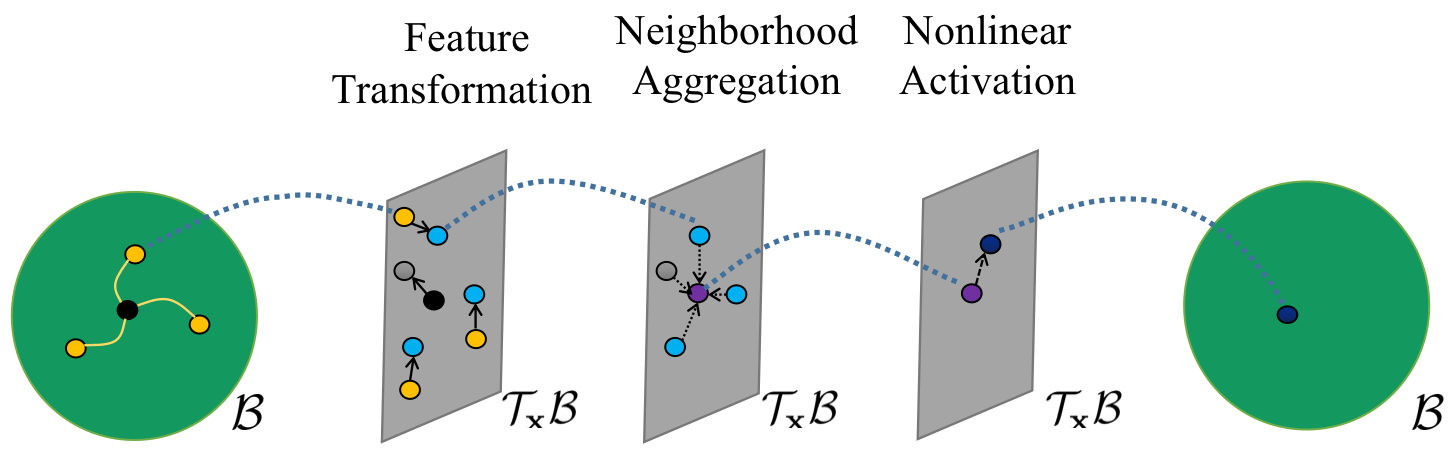} 
        }

\subfloat[Our Fully Poincar\'{e} Multi-relational GCN]{
    \includegraphics[width=0.9\columnwidth]{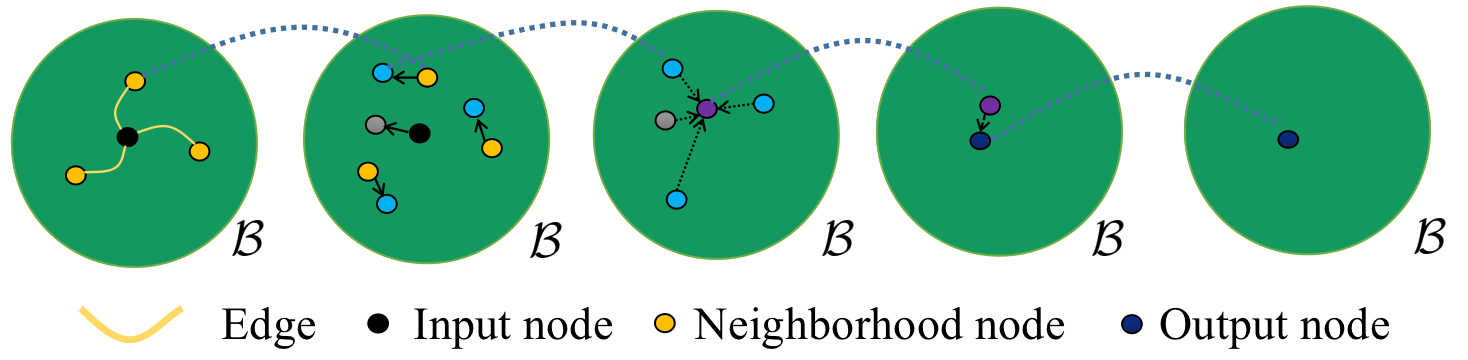}
}

\caption{(a) Classical Hyperbolic GCN, defines all operations in the tangent space, which can not maintain hyperbolic structure due to approximation. 
(b) Pur proposed FPM-GCN, is entirely based on Poincar\'{e} model without approximation.}
\label{figure_gcn}
\end{figure}

To address these limitations, in this paper, we propose two important  hyperbolic operations:

(1) Fully Poincar\'{e} Multi-relational GCN (\textbf{FPM-GCN}): which is based on Poincar\'{e} ball model and does not require tangent space approximation (as shown in Fig. \ref{figure_gcn}b). We achieve it by making two improvements --- leveraging hyperbolic isometric group transformation in feature transformation module and adopting the gyromidpoint directly in hyperbolic space instead of the tangent space. Through these two designs, FPM-GCN can maintain the hyperbolic structure and take more advantage of hyperbolic space.

(2) Hyperbolic Inner Product (\textbf{HIN}): which is a hyperbolic generalization of Euclidean inner product and exhibits more flexibility than hyperbolic distance. The derivation of HIN is theoretically sound, where we turn to M\"{o}bius gyrovector space of Poincar\'{e} ball model, make a careful analogy to the definition of Euclidean inner product, and refer to the hyperbolic law of cosines \cite{UNGAR2001135}.

With such two important proposals, we develop a \textbf{F}ully and \textbf{F}lexible \textbf{H}yperbolic \textbf{R}epresentation framework (\textbf{FFHR}) that can transfer recent Euclidean-based advances to hyperbolic space, and exploit the merits of hyperbolic space more effectively. FFHR is not a concrete model but a high-level solution blueprint. We demonstrate it by instantiating FFHR with four representative KGC methods including DistMult \cite{DBLP:journals/corr/YangYHGD14a}, ComplEx \cite{10.5555/3045390.3045609}, RESCAL \cite{10.5555/3104482.3104584} and DualE\cite{Cao_Xu_Yang_Cao_Huang_2021}. We further empirically validate their superiority over their Euclidean counterparts and hyperbolic state-of-the-arts.

In a nutshell, the main contributions of this work are summarized below:
\begin{itemize}
\item Proposing FPM-GCN, which defines all operations on Poincar\'{e} ball model. These operations do not require tangent space approximation and thus better preserve hyperbolic structure.
\item Proposing Hyperbolic Inner Product (HIN) --- a hyperbolic generalization of Euclidean inner product, which is more flexible than hyperbolic distance score function in modeling complex data patterns.
\item Developing FFHR framework that can transfer recent advanced Euclidean-based methods to hyperbolic space and instantiating it with four representative KG completion methods to justify its effectiveness.
\end{itemize}

\section{Related Work}
We have summarized the two main limitations of hyperbolic representation for KGC. In this section, we will provide a detailed review of classical Euclidean KGC methods and hyperbolic embedding methods.

\subsection{Euclidean Embedding models}
According to different similarity measures, KGC methods can be roughly divided into two categories. i) Translational methods, such as TransE \cite{10.5555/2999792.2999923}, TransH \cite{10.5555/2893873.2894046}, RotatE \cite{sun2018rotate}, measure the plausibility of facts by calculating the distance between entities. ii) Semantic matching methods, which measure similarity based on inner product score function, such as TuckER \cite{balazevic2019tucker}, QuatE \cite{NEURIPS2019_d961e9f2}, DihEdral \cite{xu-li-2019-relation}.
However, both of them require high dimensional space to achieve state-of-the-art results, which leads to high memory costs.\par
GCN-based methods have also attracted considerable research interest in recent years. R-GCN \cite{DBLP:conf/esws/SchlichtkrullKB18} firstly introduce GCN to multi-relational graph and make use of graph structure information to improve the prediction performance. CompGCN \cite{vashishth2020compositionbased} alleviates the over-parameterization problem and propose a generic framework for KGC. Due to the powerful information aggregation ability, GCN is crucial for modeling complex data patterns. \par

\subsection{Hyperbolic Embeddings}
Hyperbolic geometry is non-euclidean geometry with constant negative curvatures. \cite{10.5555/3295222.3295381}\cite{NickelK18} first embeds a homogeneous graph in hyperbolic space, which shows great improvements compared to euclidean space when embedding hierarchical data. Recently, MuRP \cite{balazevic2019multi} and AttH \cite{chami2020low} generalize hyperbolic embedding to multi-relational graph. However, they both use hyperbolic distance as the score function, which lacks flexibility and is difficult to capture complex data patterns in multi-relational graphs.\par 
Also, there is a work \cite{chamberlain2017neural} that attempts to map hyperbolic embeddings to tangent space and calculate Euclidean inner product between them as hyperbolic inner product. However, as tangent space is just a local approximation of hyperbolic space, this will inevitably cause approximation errors. In practice, it will sadly degenerate into the equivalent form with euclidean inner product.

\subsection{Hyperbolic GNN}
HNN \cite{ganea2018hyperbolic} firstly generalize the basic hyperbolic operation tools (\textit{e.g.} MLR, GRU) and formalize the hyperbolic neural network layers using the theory introduced by \cite{doi:10.1142/6625}\cite{doi:10.2200/S00175ED1V01Y200901MAS004}. Later, HNN++ \cite{shimizu2021hyperbolic} propose an improved hyperbolic MLR layer and generalize it to convolutional layer. However, the features obtained through FC layers are independent of each other, hence they don't have clear geometric meanings and can't maintain hyperbolic structure. Different from them, HGCN \cite{NEURIPS2019_0415740e} and HGNN \cite{NEURIPS2019_103303dd} simultaneously proposed hyperbolic GCN, which share the same main idea. They formalize most of their operations (feature transformation, neighborhood aggregation) in the tangent space, which causes distortion in representation learning. \par
More recently, H2HGCN \cite{Dai_2021_CVPR} propose an improved Hyperbolic-to-Hyperbolic GCN, where the operations are defined in the different isometric hyperbolic models. Their method avoids tangent space approximation but introduces undesirable transformation between different hyperbolic models, which would incur numerical instability. Also, we remark that all the above methods are defined for homogeneous graphs and do not apply to multi-relational knowledge graphs.\par
There are a few works like M$^2$GNN \cite{10.1145/3442381.3450118} and H2E \cite{9671651}  that generalize HGNN to multi-relational graph to better capture hierarchical data structures, but they define all operations in the tangent space. A more comprehensive analysis can be found in a recent survey \cite{2202.13852}. Compared to them, our FPM-GCN is defined in Poincar\'{e} model without any approximation error. Also, our FPM-GCN does not need transformation between different isometric models in H2HGCN \cite{Dai_2021_CVPR}, which is more stable and expressive.

\section{Preliminaries}

In this section, we first propose the formal definition of knowledge graph completion task and then introduce some background knowledge related to hyperbolic geometry. 

\subsection{Task Formulation.}
 Let $\mathcal{G}=\{\mathcal{E}, \mathcal{R}, \mathcal{L}\}$ be an instance of KG, where $\mathcal{E}$, $\mathcal{R}$ and $\mathcal{L}$ represent the sets of entities, relations and facts, respectively.  Each fact $f \in \mathcal{L}$ is represented as a triple $(h,r,t)\in \mathcal{E}\times \mathcal{R} \times \mathcal{E}$, describing there is a relationship $r \in \mathcal{R}$ from the head entity $h$ to the tail entity $t$. The task of knowledge graph completion involves inferring the missing edges in the knowledge graph. It aims at learning a function $s:\mathcal{E}\times\mathcal{R}\times\mathcal{E}\to\mathbb{R}$ that assigns a higher score for valid triplets than the invalid ones.

\subsection{Riemannian Manifold.} An n-dimensional manifold $\mathcal{M}$ is a topological space where for each point $\mathbf{x} \in \mathcal{M}$ its neighborhood can be approximated locally by tangent space $\mathcal{T}_\mathbf{x} \mathcal{M}$. A manifold $\mathcal{M}$ equipped with a metric tensor $g_\mathbf{x}$ is a \textit{Riemannian manifold}, denoted by $(\mathcal{M}, g)$. The metric tensor $g_\mathbf{x}: \mathcal{T}_\mathbf{x} \mathcal{M} \times \mathcal{T}_\mathbf{x} \mathcal{M} \rightarrow \mathbb{R}$ provides local information regarding the angle and length of the tangent vectors in $\mathcal{T}_\mathbf{x}\mathcal{M}$, which induce the global distance through integration. Global distance is defined as $d(\mathbf{x}, \mathbf{y})= \int_0^1 \sqrt{g_{\gamma(t)}( \dot{\gamma}(t), \dot{\gamma}(t))} \mathrm{d}t $, where smooth path $\gamma$ is \textit{geodesic} such that $\gamma(0) = \mathbf{x}$ and $\gamma(1)=\mathbf{y}$. The \textit{Exponential map} $\exp_\mathbf{x}(\mathbf{v}) : \mathcal{T}_\mathbf{x} \mathcal{M} \rightarrow \mathcal{M}$ map tangent vector $\mathbf{v} \in \mathcal{T}_\mathbf{x} \mathcal{M}$ to the manifold along the geodesic. Inversely, the \textit{Logarithmic map} $\log_\mathbf{x}(\mathbf{y}): \mathcal{M} \rightarrow \mathcal{T}_\mathbf{x} \mathcal{M}$ project $\mathbf{y} \in \mathcal{M}$ back to the tangent space, naturally satisfying $\log_\mathbf{x}(\exp_\mathbf{x}(\mathbf{v}))=\mathbf{v}$.

\subsection{Hyperbolic space and Poincar\'{e} Ball Model.} Hyperbolic space is a specific Riemannian manifold that has constant negative curvature. Among five different isometric models for hyperbolic space, we choose Poincar\'{e} ball model for derivation as it supports some important vector operations \cite{book1}\cite{book2}, such as M\"{o}bius addition, M\"{o}bius matrix-vector multiplication. These vector operations help to formalize graph neural network in hyperbolic space. Formally, let $(\mathcal{B}_c^n, g^c)$ be a Poincar\'{e} ball model with constant curvature $-c$.  The Riemannian metric is:
\begin{equation}
g_\mathbf{x}^c = (\lambda_\mathbf{x}^c)^2 \mathbf{I}_n, \quad \lambda_\mathbf{x}^c = \frac{2}{1-c||\mathbf{x}||^2} \label{con:poincare_defination}
\end{equation}
The M\"{o}bius addition and M\"{o}bius matrix-vector multiplication are shown as follows:
\begin{equation}
    \mathbf{x} \oplus_c \mathbf{y} = \frac{(1+2c\left< \mathbf{x}, \mathbf{y} \right> +c||\mathbf{y}||^2)\mathbf{x} + (1-c||\mathbf{x}||^2)\mathbf{y}}{1 + 2c\left< \mathbf{x}, \mathbf{y} \right> + c^2||\mathbf{x}||^2||\mathbf{y}||^2},
\end{equation} \par
\begin{equation}
    \mathbf{W} \otimes_c \mathbf{x} = exp_\mathbf{0}^c(\mathbf{W}\cdot log_\mathbf{0}^c(\mathbf{x})),
\end{equation}
where $exp_\mathbf{0}^c$ and $log_\mathbf{0}^c$ are Exponential
map and Logarithmic map for $(\mathcal{B}_c^n, g^c)$, and $\mathbf{0}$ is the origin of Poincar\'{e} ball model.
Using the M\"{o}bius addition, we can rewrite some important formulas with clearer geometric interpretation. The geodesic connecting points $\mathbf{x}, \mathbf{y} \in \mathcal{B}_c$ is given by: 
\begin{equation*}
    \gamma_{\mathbf{x} \rightarrow \mathbf{y}}(t) := \mathbf{x} \oplus_c (-\mathbf{x}\oplus_c \mathbf{y}) \otimes_c t ,
\end{equation*}
\begin{equation*}
    s.t. \gamma_{\mathbf{x} \rightarrow \mathbf{y}}(0) = \mathbf{x} \  and \  \gamma_{\mathbf{x} \rightarrow \mathbf{y}}(1) = \mathbf{y}.
\end{equation*}
The close-form derivation of exponential and logarithmic maps are:
\begin{equation*}
    exp_\mathbf{x}^c(\mathbf{v}) = \mathbf{x} \oplus_c \left( tanh\left( \sqrt{c} \frac{\lambda_\mathbf{x}^c ||\mathbf{v}||}{2} \right) \frac{\mathbf{v}}{\sqrt{c}||\mathbf{v}||} \right),
\end{equation*}
\begin{equation*}
    log_\mathbf{x}^c(\mathbf{y}) = \frac{2}{\sqrt{c} \lambda_\mathbf{x}^c}artanh(\sqrt{c}||-\mathbf{x} \oplus_c \mathbf{y}||)\frac{-\mathbf{x}\oplus_c \mathbf{y}}{||-\mathbf{x}\oplus_c\mathbf{y}||},
\end{equation*}

where $\mathbf{x} \in \mathcal{B}_c$, $\mathbf{v} \in \mathcal{T}_\mathbf{x}\mathcal{B}_c$, and $\lambda_\mathbf{x}^c$ is the conformal factor of $\mathcal{B}_c$. If we observe these formulas carefully, we will find that they are very similar to their Euclidean counterpart. Different from Euclidean addition, the M\"{o}bius addition is a non-commutative and non-associative operation. 

Also, the hyperbolic distance function is defined as:
\begin{equation}
    d^{\mathcal{B}_c}(\mathbf{x},\mathbf{y})=\frac{2}{\sqrt{c}}arctanh(\sqrt{c}||-\mathbf{x}\oplus_c \mathbf{y}||).
    \label{equ_distance}
\end{equation}

With the help of exp and log mapping functions, previous work propose the classical feature transformation and neighborhood aggregation modules, which are defined in the tangent space of the origin. \par
If you are looking for further knowledge about Poincar\'{e} Ball Model, you can refer to \cite{doi:10.2200/S00175ED1V01Y200901MAS004}.  \par

\subsection{Graph  Convolutional Networks}
Given the graph $\mathcal{G} = (\mathcal{E}, \mathcal{R})$ with entity set $\mathcal{E}$ and relation set $\mathcal{R}$, we can formulate general euclidean graph  convolutional network into two main steps. Let $x_{i}^{l}$ denotes the $i$-th entity's euclidean representations at the $l$-th layer. 

\subsubsection{feature transformation}
\begin{equation}
    \mathbf{m}_i^{l, E} = \mathbf{W}_r^l \mathbf{x}_i^{l-1, E} + \mathbf{b}_r^l
\end{equation}
where $\mathbf{W}_r^l$ denotes the transform matrix of relation r and $\mathbf{b}_r$ denotes the bias vector of relation r.

\subsubsection{neighborhood aggregation}
\begin{equation}
    \mathbf{x}_i^{l,E} = \sigma ( \mathbf{m}_i^{l,E} + \sum_{j\in N(i)}  \mathbf{m}_{j}^{l,E} )
\end{equation}
where $N(i)$ denotes the set of neighborhood of entity i for its outgoing edges. $\sigma(\cdot)$ denotes activation function. \par
Then the message of neighborhood can be received through stacking multiple graph convolutional layers, which is useful for downstream tasks.

\section{Our Proposals}
In this section, we first detail the definition of FPM-GCN and HIN. Then we present our FFHR framework and show how it generalizes Euclidean-based advances to hyperbolic space. Finally, we instantiate FFHR with four representative KGC methods.

%
%
%
\subsection{Fully Poincar\'{e} Multi-relational GCN (FPM-GCN)}
In this section, we propose a novel hyperbolic GCN, which does not require the approximation via the tangent space nor the transformation between different isometric models. As shown in Fig \ref{figure_gcn}, FPM-GCN entirely operates on hyperbolic space, with three important modules:  \par
\subsubsection{Feature Transformation Module.} From the work \cite{57ffe2472e1a482a86a80b723bfe02d8}\cite{tang-etal-2020-orthogonal}, we know that isometric group transformation (orthogonal transformation in Euclidean space) is more expressive and able to stabilize neural network. Hence, we leverage isometric group transformation in feature transformation module. Concretely, we formalize the feature transformation module as the composite of a left gyrotranslation and orthogonal transformation as:
\begin{equation}
    \mathbf{m}_i^{l,\mathcal{B}_c} = \mathbf{b}_r^l \oplus_c \mathbf{W}_r^l \mathbf{x}_i^{l-1,\mathcal{B}_c}, \quad s.t. (\mathbf{W}_r^{l})^T\mathbf{W}_r^l =\mathbf{I}, \label{con:relation_transform_module}
\end{equation}
where $\mathbf{W}_r^l$ are relation-aware transformation matrices and $\mathbf{b}_r^l$ are relation-aware transformation vectors. We show that the proposed module satisfies the following two desirable properties:
\begin{prop} \textbf{Manifold Preserving.} \par
The Feature transformation module is manifold preserving. The transformed vectors still lie on the Poincar\'{e} ball model.
\label{manifold}
\end{prop}

\begin{proof} 
Firstly, we divide our feature transformation module into two parts. Orthogonal transformation part: 
\begin{equation*}
    \Phi(\mathbf{x}) = \mathbf{W}\cdot \mathbf{x},\ s.t.\ \mathbf{W}^T\mathbf{W} =\mathbf{I}
\end{equation*}
M\"{o}bius transformation part:
\begin{equation*}
    \mathbf{L}_\mathbf{u}(\mathbf{x}) = \mathbf{u} \oplus_c \mathbf{x} 
\end{equation*}
Easily, we have $||\Phi(\mathbf{x})|| = ||\sqrt{\mathbf{x}^T\mathbf{W}^T\mathbf{W}\mathbf{x}}||= ||\mathbf{x}||<\frac{1}{c}$ for the first part. For the M\"{o}bius transformation part, we want to proof that, 
\begin{equation*}
    ||\mathbf{u}\oplus_c \mathbf{x}|| < \frac{1}{\sqrt{c}}
\end{equation*}
which is equal to, 
\begin{equation*}
    (\mathbf{u} \oplus \mathbf{x})^T(\mathbf{u} \oplus \mathbf{x}) < \frac{1}{c}.
\end{equation*}
Using the M\"{o}bius addition formula, we just need to proof:
\begin{equation*}
    c(||\mathbf{u}||^2 - \frac{1}{c})(||\mathbf{x}||^2 - \frac{1}{c})(1 + 2c\left< \mathbf{u}, \mathbf{x} \right> + c^2||\mathbf{u}||^2||\mathbf{x}||^2) > 0
\end{equation*}
We already have $||\mathbf{u}||^2<\frac{1}{c}$, $||\mathbf{x}||^2< \frac{1}{c}$. And using Cauchy-Schiwarz inequality, easily we have $1 + 2c\left< \mathbf{u}, \mathbf{x} \right> + c^2||\mathbf{u}||^2||\mathbf{x}||^2 > 0$. Hence, the above inequality is satisfied, and the two transformation part are both manifold preserving. This completes our proof.

\end{proof} 

\textbf{Remark} This proposition ensures that our feature transformation entirely operates in the Poincar\'{e} Ball without approximation. 
\begin{theorem} \textbf{Expressive.}\par
Our feature transformation defines an isometry of $\mathcal{B}_c$ with respect to $d_{\mathcal{B}_c}$: $$d_{\mathcal{B}_c}(Transformed(\mathbf{x}),Transformed(\mathbf{y})) = d_{\mathcal{B}_c} (\mathbf{x}, \mathbf{y}),$$ and every isometry of $(\mathcal{B}_c, d_{\mathcal{B}_c}) $ can be expressed in the same form as our feature transformation. 
\label{expressive}
\end{theorem}

\begin{proof}
Firstly, we proof our feature transformation is an isometry of $\mathcal{B}_c$ respect to $d_{\mathcal{B}_c}$. We have the distance function: 
\begin{equation*}
     d^{\mathcal{B}_c}(\mathbf{x},\mathbf{y})=\frac{2}{\sqrt{c}}arctanh(\sqrt{c}||-\mathbf{x}\oplus_c \mathbf{y}||)
\end{equation*}
Hence we just need to proof $||-\mathbf{L}_\mathbf{u}(\mathbf{x}) \oplus \mathbf{L}_\mathbf{u}(\mathbf{y})|| = ||-\mathbf{x}\oplus \mathbf{y}||, \ ||-\Phi(\mathbf{x})\oplus \Phi(\mathbf{y})|| =||-\mathbf{x}\oplus \mathbf{y}||, \ \forall \mathbf{x},\mathbf{y} \in \mathcal{B}$. \\
For the first identity, we have
\begin{equation*}
\begin{split}
    &||-\mathbf{L}_\mathbf{u}(\mathbf{x}) \oplus \mathbf{L}_\mathbf{u}(\mathbf{y})|| \\ 
    =& ||-(\mathbf{u}\oplus\mathbf{x})\oplus(\mathbf{u}\oplus\mathbf{y})|| \\ 
    =& ||-(\mathbf{u}\oplus\mathbf{x})\oplus\left( \mathbf{u}\oplus (\mathbf{x}\oplus(-\mathbf{x}\oplus\mathbf{y}))\right)|| \\
    =& ||-(\mathbf{u}\oplus\mathbf{x})\oplus (\mathbf{u}\oplus\mathbf{x})\oplus gyr[\mathbf{u}, \mathbf{x}](-\mathbf{x} \oplus \mathbf{y}) ||\\
    =& ||gyr[\mathbf{u}, \mathbf{x}](-\mathbf{x} \oplus \mathbf{y}) ||\\
    =&||-\mathbf{x}\oplus \mathbf{y}||
\end{split}
\end{equation*}
In the derivation process, we use the conclusion mentioned in Lemma 1.\\
For the second identity, we have
\begin{equation*}
    \begin{split}
        &||-\Phi(x) \oplus \Phi(y)|| \\
        =&||\mathbf{W}\frac{(1+2c\left< -\mathbf{x}, \mathbf{y} \right> +c||\mathbf{y}||^2)(-\mathbf{x}) + (1-c||\mathbf{x}||^2)\mathbf{y}}{1 + 2c\left< -\mathbf{x}, \mathbf{y} \right> + c^2||\mathbf{x}||^2||\mathbf{y}||^2}|| \\
        =&||-\mathbf{x}\oplus \mathbf{y}||.
    \end{split}
\end{equation*}\par

Secondly, we argue that every isometry of $(\mathcal{B}_c, d_{\mathcal{B}_c}) $ can be expressed in the same form as our feature transformation. We denote isometry group of $(\mathcal{B}_c, d_{\mathcal{B}_c}) $ as $Iso(\mathcal{B}_c, d_{\mathcal{B}_c})$. \par
Supposing that $\mathbf{R} \in Iso(\mathcal{B}_c, d_{\mathcal{B}_c})$, by definition, $\mathbf{R}$ is a bijective self-map of $\mathcal{B}$. Using theorem 11 shown in \cite{Suk}, we have $\mathbf{R}=\mathbf{L}_{\mathbf{R}(\mathbf{0})} \circ \tau$, where $\tau$ is a bijective self-map of $\mathcal{B}$ fixing $\mathbf{0}$. Following the proof of theorem 18(2) in \cite{book_Rassias}, $\mathbf{L}_{\mathbf{R}(\mathbf{0})}^{-1} = \mathbf{L}_{-\mathbf{R}(\mathbf{0})}$. Hence, we have $\tau = \mathbf{L}_{-\mathbf{R}(\mathbf{0})} \circ \mathbf{R}$ and $\tau$ is an isometry of $(\mathcal{B}_c, d_{\mathcal{B}_c})$. We can find that $\tau$ preserves the M\"{o}bius gyrometric. Using the theorem 3.2 in \cite{article_Abe}, we observe the $\tau$ is a bijective orthogonal transformation on $\mathcal{R}^n$. This completes our proof.
\end{proof}
\textbf{Remark} We notice that other feature transformation modules in previous HGCN are just special cases of isometric group transformation, which causes poor expressive ability and performance. \par
 For multi-relational graph, general orthogonal matrix is high memory-intensive and the optimization is also time-consuming \cite{57ffe2472e1a482a86a80b723bfe02d8}. Hence in practice, we formalize the matrix as a block diagonal matrix:
 \begin{equation}
     \mathbf{W}_r = diag(A(\theta_{1,r}), A(\theta_{2,r}), \cdots, A(\theta_{\frac{n}{2}, r})),
\end{equation}
\begin{equation}
     A(\theta_{i,r}) =  \begin{bmatrix}
   cos(\theta_{i,r}) & -sin(\theta_{i,r}) \\
   sin(\theta_{i,r}) & cos(\theta_{i,r})
  \end{bmatrix}.
 \end{equation}

\subsubsection{Neighborhood Aggregation Module.}
Here we propose to use the M\"{o}bius gyromidpoint \cite{book2}\cite{shimizu2021hyperbolic} to calculate the weighted center point of the neighborhoods, which can be directly calculated based on Poincar\'{e} ball model without any approximation or space transformation. The center point can be characterized as a minimizer of the weighted sum of calibrated squared distance.
\begin{equation}
   \mathbf{x}_i^{k,l,\mathcal{B}_c} = \frac{1}{2} \otimes_c \left( \frac{\sum_{j\in N(i)} v_{i,j}^k \cdot \lambda_{\mathbf{m}_j^{l,\mathcal{B}_c}}^c \cdot \mathbf{m}_j^{l,\mathcal{B}_c}}{\sum_{j \in N(i)} |v_{i,j}^k|\cdot (\lambda_{\mathbf{m}_j^{l,\mathcal{B}_c}}^c -1)}\right),
\end{equation}
where $v_{i,j}^k$ is k-th attention weight and $N(i)$ is the set of one-hop neighborhoods of $i$. $ \lambda_{\mathbf{m}_j^{\mathcal{B}_c}}^c$ is the conformal factor defined in Eq. \ref{con:poincare_defination}. \par
To get the attention weights, we take attention module:
\begin{equation}
    v_{i,j}^k = LeakyReLU((\mathbf{a}_h^k)^T \mathbf{m}_i^{l,\mathcal{B}_c} + (\mathbf{a}_t^k)^T \mathbf{m}_j^{l,\mathcal{B}_c}),
\end{equation}
where $\mathbf{a}_h$ and $\mathbf{a}_t$ are attention vectors. It's noticeable that the attention weights do not contain structural information and thus can be obtained directly. \par
Furthermore, multi-head attention can be employed, if we treat them as points with equal weights and use M\"{o}bius gyromidpoint to aggregate them again. 

\subsubsection{Nonlinear Activation Module.} Interestingly, performing a common nonlinear activation function (\eg ReLU, Softmax) on the Poincar\'{e} ball Model is manifold preserving. Here we formalize the nonlinear activation module in FPM-GCN as:
\begin{equation}
    \mathbf{x}_i^{l,\mathcal{B}_c} = \frac{1}{\sqrt{c}}\cdot \sigma(\sqrt{c}\cdot \mathbf{x}_i^{l,\mathcal{B}_c}),
\end{equation} \par
We simply chose the LeakyReLU for implementation.

\subsection{Hyperbolic Inner Product (HIN)}
Before going deep into our HIN, we would like to summarize the recent advances in Euclidean space and conduct a brief analysis of the score functions, which both provide useful insights into their characteristics.




\begin{figure}
	\centering
	\includegraphics[width=1.0\columnwidth]{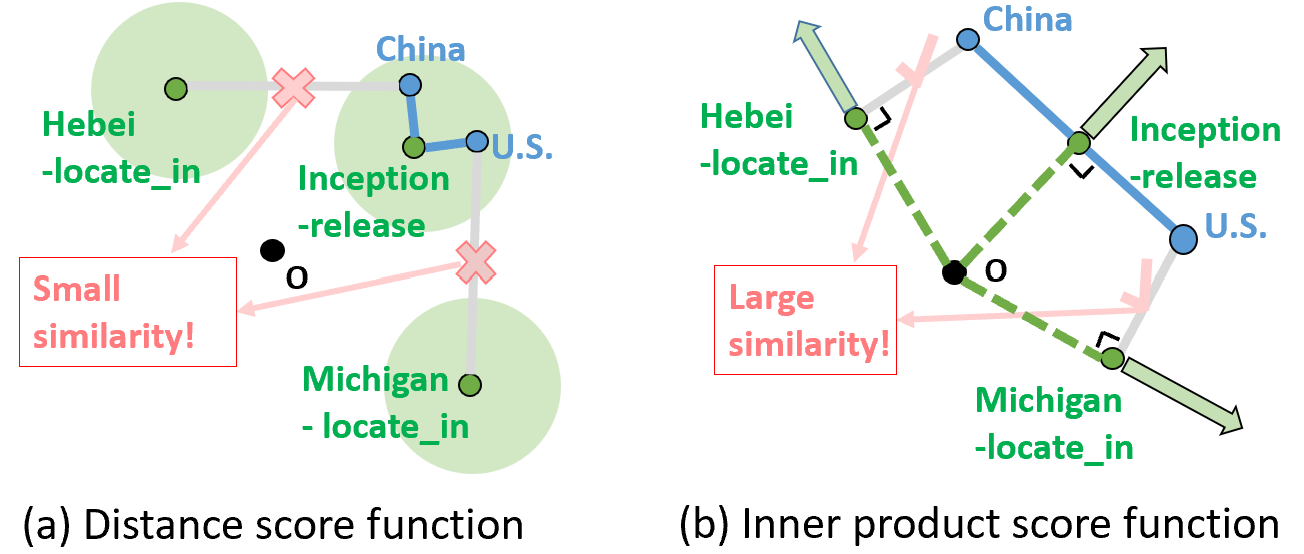}
	\caption{``Inception'' ``release'' in both ``U.S.'' and ``China''. IP just constrains the projection of an embedding vector on a specific direction, hence it can maintain the difference between "U.S." and "China". On the contrary, distance function constrains a vector in all dimensions and fall short in maintaining difference and flexibility.}
	\label{figure_cos}
\end{figure}

\subsubsection{Comparions of Two Score Functions.} 
Formally, the two types of score functions can be written as follow:
\begin{equation}
    s_{dist}(h,r,t) = -||\phi_r(\mathbf{e}_{h}) - \mathbf{e}_{t}||^2,
\end{equation}
\begin{equation}
    s_{inner}(h,r,t) = \left< \phi_r(\mathbf{e}_{h}), \mathbf{e}_{t} \right>,
\end{equation} \par
where $\mathbf{e}_{(\cdot)}$ denotes the learned embedding for each entity and $\phi_r(\mathbf{e}_{h})$ is transformed embedding of the head entity $h$ after the relation aware transformation $\phi_r(\cdot)$. The score functions resort to the similarity (inner product or distance) between $\phi_r(\mathbf{e}_{h})$ and $\mathbf{e}_{t}$ to estimate the plausibility of a given fact.

Note that complex relation patterns are common in a KG. Typically, take Fig. \ref{figure_toy} as an example, there exist complex many-to-many relations, \ie a head entity may connect various tail entities with the same relation. For example, the ``Inception'' ``release'' in both ``U.S.'' and ``China''. When learning a representation to capture such complex relation patterns, distance-based score function would fall short. To better understand this point, we give a toy example of the learned representation using two score functions as illustrated in Fig. \ref{figure_cos}. As can be seen, the distance-based function would give quite a strict constraint --- it forces the embeddings of the tail entities to be absolutely close if they are connected by a common head entity. It is inflexible and may incur contradiction. For example, "China" and "U.S." shall also be close to their other exclusive tail entities, such as "Hebei-locate$\_$in" and "Michigan-locate$\_$in" respectively. Such conditions can not be achieved simultaneously --- making distinctive entities "China" and "U.S." closer would inevitably destroy the
inherent differences between them. For inner product, this negative effect can be mitigated, as inner product is a relatively mild measure --- it only constrains the similarity of the entities in a specific direction. Inner product is flexible and can maintain differences for complex data patterns. e.g, "China" can share higher similarities with both "Inception-release" and "Hebei-local$\_$in".  



Although our analysis is intuitive and not rigorous, it is useful to understand how the inner product is superior over the distance. In what follows, we also conduct an empirical study to validate this point.

\subsubsection{Summary of Recent Euclidean Advances} 
\begin{figure}
	\centering
	\includegraphics[width=0.7\columnwidth]{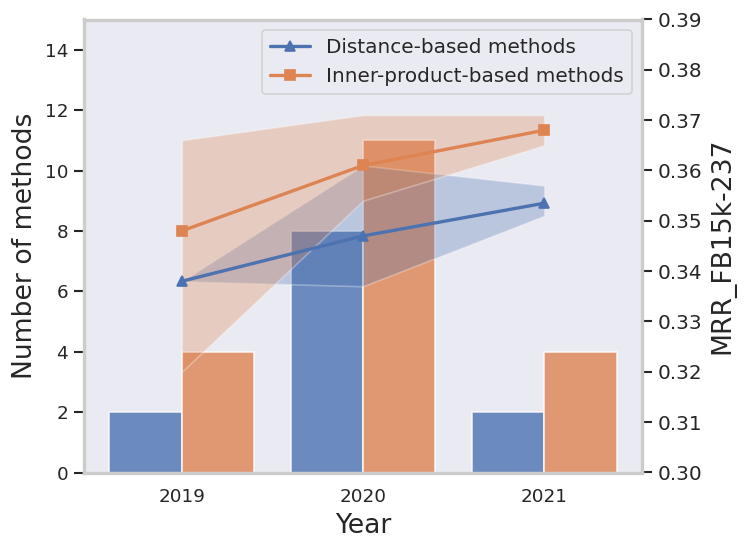}
	\caption{The bar chart and the line chart present the number of KGC methods and their average MRR results on a benchmark in recent years respectively. (The shaded region represents the range of MRR in different years.)}
	\label{figure_summary}
\end{figure}

To make a comparison of two score functions, we conduct a comprehensive survey of recent Eudiean-based methods for KGC published in top conferences SIGKDD, NIPS, WWW, etc.  Fig. \ref{figure_summary} illustrates the number and performance comparisons of the methods with distance-based score function or inner-product-based function. We could find that inner-product score function attracts more attention in recent years. And when embedding complex data patterns (\textit{e.g.} FB15K-237), inner-product-based methods perform better on average. Beyond that, state-of-the-art results are also achieved by inner-product-based methods. These statistical results may suggest the superiority of inner product score function for embedding complex data patterns. \par 
Above theoretical analysis and statistical results motivate us to explore the inner product in hyperbolic space.

\subsubsection{Inner Product in Hyperbolic Space} Although inner product is superior in KGC, it still lacks good definition in hyperbolic space. To calculate inner product in hyperbolic space, current work \cite{chamberlain2017neural} generally maps hyperbolic embeddings to tangent space and calculates Euclidean inner product between them. Concretely, this can be formulated as:
\begin{equation}
    \left< \mathbf{x}, \mathbf{y} \right>_{\mathcal{B}_c^n} = \left< log_{\mathbf{0}}^c(\mathbf{x}), log_{\mathbf{0}}^c(\mathbf{y}) \right>
\end{equation}
However, this formulation has two main drawbacks: 1) Tangent space is just a local approximation of hyperbolic space, and this will inevitably cause approximation errors. Using prevailing optimization methods in \cite{chami2020low}, it will sadly degenerate into the equivalent form with euclidean inner product. 2) The formulation does not obey the hyperbolic law of cosines and thus can not maintain the hyperbolic structure of data.\par
\begin{figure}
	\centering
    \includegraphics[width=0.9\columnwidth]{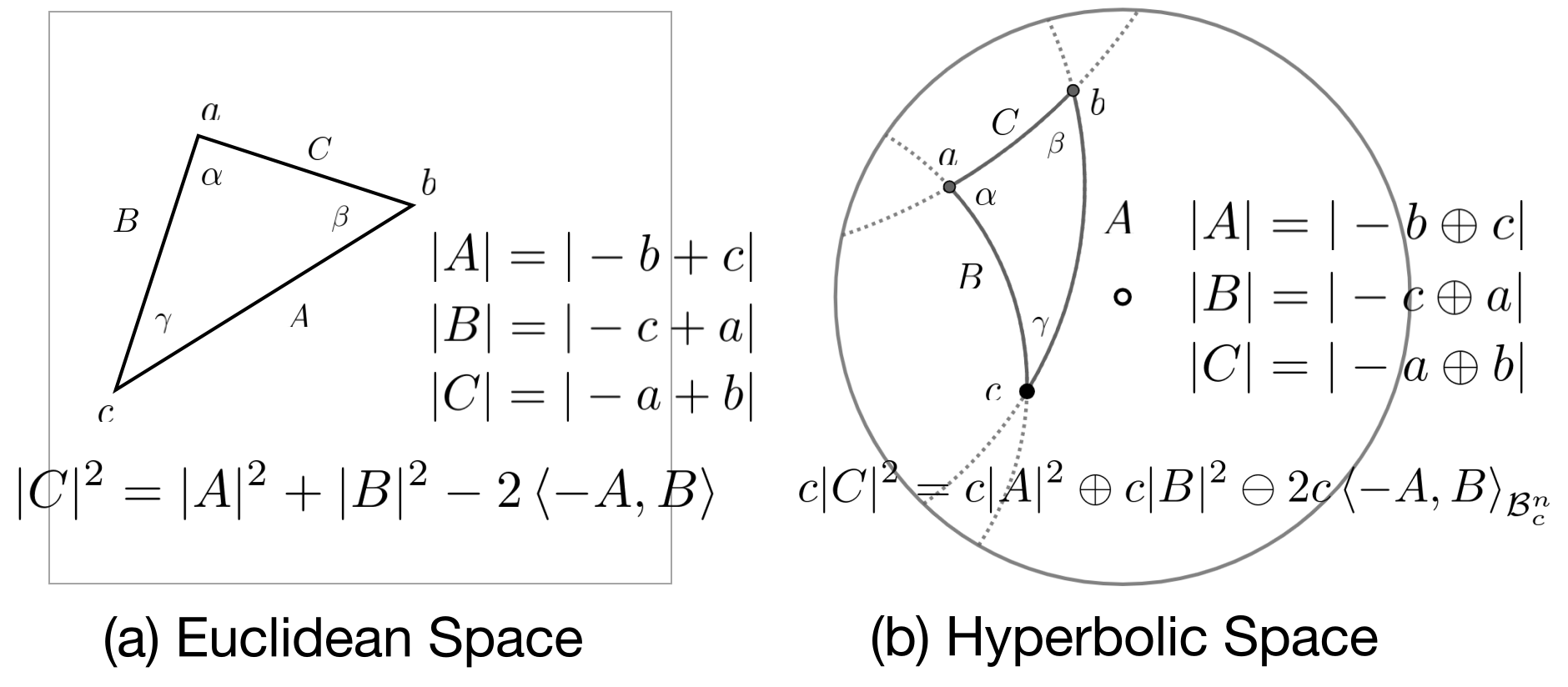}
	\caption{Illustration of (a) Euclidean laws of cosines and (b) Hyperbolic laws of cosines on Poincar\'{e} ball model.}
	\label{figure_inner}
\end{figure}

To solve this problem, in this subsection, we present a meaningful and useful generalization of Euclidean inner product to hyperbolic space. We achieve it by referring to the definition of Euclidean inner product and the hyperbolic law of cosines \cite{UNGAR2001135}. Formally, for any hyperbolic triangle $\bigtriangleup \mathbf{0} \mathbf{x} \mathbf{y}, \mathbf{x},\mathbf{y} \in \mathbb{B}_c^n $, we have:
\begin{equation*}
    c|| \mathbf{C} ||^2  = c ||\mathbf{A}||^2 \oplus_c c ||\mathbf{B} ||^2 \ominus_c 2c\Gamma(\mathbf{A}, \mathbf{B}, \psi),
\end{equation*}
where $\mathbf{A}=-\mathbf{0}\oplus_c \mathbf{x}$, $ \mathbf{B}=-\mathbf{0} \oplus_c \mathbf{y}$, $\mathbf{C}=-\mathbf{x} \oplus_c \mathbf{y}$, $\psi$ is the hyperbolic angle at origin and $\Gamma(\mathbf{A}, \mathbf{B}, \psi)$ denotes the specific similarity function between $\mathbf{A}$ and $\mathbf{B}$. Here hyperbolic distance $||\mathbf{C}||^2$ can be decomposed into the two hyperbolic norms $||\mathbf{A}||^2$ and $||\mathbf{B}||^2$ M\"{o}bius minus the similarity function $\Gamma(\mathbf{A}, \mathbf{B}, \psi)$. As shown in Fig. \ref{figure_inner}, analogy to the definition of Euclidean inner product (according to the Law of cosines), we denote the similarity function $\Gamma(\mathbf{A}, \mathbf{B}, \psi)$ as ``Hyperbolic Inner Product'', which is:
\begin{equation*}
\begin{split}
    &\left< \mathbf{x}, \mathbf{y} \right>_{\mathcal{B}_c^n} = \Gamma(\mathbf{A}, \mathbf{B}, \psi) \\
    &=\frac{  ||\mathbf{A} || \cdot ||\mathbf{B}||\cdot cos\psi}{(1+c^2||\mathbf{A} ||^2)(1+c^2||\mathbf{B}||^2) - 2c^2||\mathbf{A} ||\cdot||\mathbf{B}||\cdot cos\psi},
\end{split}
\end{equation*}
where $\mathbf{A}=-\mathbf{0}\oplus_c \mathbf{x}$, $ \mathbf{B}=-\mathbf{0} \oplus_c \mathbf{y}$, and $\psi$ is the hyperbolic angle at origin. Easily, we can observe that $\lim_{c\rightarrow 0}\left< \mathbf{x}, \mathbf{y} \right>_{\mathcal{B}_c^n}= \left< \mathbf{x}, \mathbf{y} \right>$, indicating that our hyperbolic inner product is a generalization of Euclidean inner product thus inherit the flexibility of inner-product-based score function. Our HIN is also theoretically sound, as we just make an analogy of the Euclidean inner product and resort to hyperbolic law of cosines, without making any additional assumptions. \par

\begin{table*}[!t]
\renewcommand{\arraystretch}{1.9}
    \centering
    \caption{A summary of how our FFHR generalize Euclidean methods to hyperbolic space.}
    \begin{tabular}{c c c}
    \hline
    Module   & Euclidean space &  FFHR \\
    \hline
    Feature transformation &  $\mathbf{m}_i^{l, E} = \mathbf{W}_r^l \mathbf{x}_i^{l-1, E} + \mathbf{b}_r^l$ & $\mathbf{m}_i^{l,\mathcal{B}_c} = \mathbf{b}_r^l \oplus_c \mathbf{W}_r^l \mathbf{x}_i^{l-1,\mathcal{B}_c}$  \\
    \hline
    Neighborhood aggregation & $ \mathbf{x}_i^{l,E} = \sum_{j\in N(i)} v_{i,j}^k \cdot \mathbf{m}_{j}^{l,E}$  & $\mathbf{x}_i^{k,l,\mathcal{B}_c} = \frac{1}{2} \otimes_c \left( \frac{\sum_{j\in N(i)} v_{i,j}^k \cdot \lambda_{\mathbf{m}_j^{l,\mathcal{B}_c}}^c \cdot \mathbf{m}_j^{l,\mathcal{B}_c}}{\sum_{j \in N(i)} |v_{i,j}^k|\cdot (\lambda_{\mathbf{m}_j^{l,\mathcal{B}_c}}^c -1)}\right)$\\
    \hline
    Nonlinear activation & $\mathbf{x}_i^{l,E} = \sigma (\mathbf{x}_i^{l,E})$  &  $\mathbf{x}_i^{l,\mathcal{B}_c} = \frac{1}{\sqrt{c}}\cdot \sigma(\sqrt{c}\cdot \mathbf{x}_i^{l,\mathcal{B}_c})$ \\
    \hline
    Inner product score function & $\left< \phi_r(\mathbf{e}_h), \mathbf{e}_t \right>$  & $\left< \phi_r(\mathbf{e}_h), \mathbf{e}_t \right>_{\mathcal{B}_c}$ \\
    \hline
    \end{tabular}
    \label{tab:my_label}
\end{table*}

\subsection{Fully and Flexible Hyperbolic Representation framework (FFHR)}
With above two important hyperbolic improvements, we develop a fully and flexible hyperbolic representation framework FFHR. FFHR is not a concrete model but a high-level solution blueprint, which is able to transfer recent Euclidean-based advances to hyperbolic space. As shown in Table \ref{tab:my_label}, we demonstrate how FFHR framework generalizes Euclidean inner-product-based methods to hyperbolic space. \par
For KGC methods, using general ternary scoring function $\mathcal{F}(h, r, t)$ is time-consuming and inefficient. Hence mostly they can be expressed as $\mathcal{F}(\phi_r(h), t)$. According to different score functions $\mathcal{F}$, these methods can be divided into two categories, distance-based methods, and inner-product-based methods. As converting distance-based methods to hyperbolic space is relatively easy and has been studied by previous work, then we mainly discuss how our FFHR framework transfer all inner-product-based methods to hyperbolic spaces. \par 
The difference between inner-product-based methods is their transformation functions $\phi_r$.
To transfer these models, maintaining their transformation functions $\phi$, we replace inner product score function with our HIN function and replace their Euclidean GCN with our FPM-GCN. In this paper, we simply instantiate FFHR with four representative KGC methods including DistMult \cite{DBLP:journals/corr/YangYHGD14a}, ComplEx \cite{10.5555/3045390.3045609}, RESCAL \cite{10.5555/3104482.3104584} and DualE\cite{Cao_Xu_Yang_Cao_Huang_2021} and test their performance in our experiments. 
\begin{equation}
    s(h,r,t)= \left< \mathbf{M}_r \otimes_c \mathbf{e}_h, \mathbf{e}_t \right>_{\mathcal{B}_c^n},
\end{equation}
where $\mathbf{e}_{(.)}\in \mathcal{B}_c^n$ denotes entity embddding and $\mathbf{M}_r$ denotes relation transformation matrix. 
\begin{itemize}
    \item \textbf{DistMult-FFHR}: $\mathbf{M}_r$ is a diagonal matrix;
    \item \textbf{ComplEx-FFHR}: $\mathbf{M}_r$ are block-diagonal matrices;
    \item \textbf{DualE-FFHR}: $\mathbf{M}_r$ are block-diagonal matrices;
    \item \textbf{RESCAL-FFHR}: $\mathbf{M}_r$ is a general matrix.
\end{itemize}
We also test their variants without FPM-GCN, denoted as DistMult-H, ComplEx-H, RESCAL-H, and DualE-H.

\begin{table}[t]
\centering
\renewcommand{\arraystretch}{1.3}
\caption{Statistics of benchmark datasets.}
\begin{tabular}{l c c c} 
\hline
Dataset &\#Entity &\#Relation &\#Triples\\
\hline
WN18RR & 40,943 &11 &93,003 \\
FB15k-237 &14,541 &237 &310,116\\
WN18   & 40,943 &18 & 151,442   \\
\hline
\end{tabular}
\label{table_dataset}
\end{table}

\begin{table*}[!t]
\centering
\renewcommand{\arraystretch}{1.3}
\caption{KGC Performance comparison in low dimension case (d=32), where ``IP'', "HS", "GCN" denote whether using Inner product, Hyperbolic Space, GCN respectively. The best results are in bold and the second best are underlined. }
\resizebox{\textwidth}{!}{
\begin{tabular}{l c c c c c c c c c c c c c c c c c c}
\hline
  &\multicolumn{3}{c}{characteristic} &   &\multicolumn{4}{c}{WN18RR} & &\multicolumn{4}{c}{FB15K-237}  & &\multicolumn{4}{c}{WN18}\\
\cline{2-4} \cline{6-9} \cline{11-14} \cline{16-19} 
 model &IP? &HS? &GCN? & &MRR &H@1 &H@3 &H@10 &  &MRR &H@1 &H@3 &H@10 & &MRR &H@1 &H@3 &H@10 \\
\hline
RotatE &$\times$ &$\times$ &$\times$ & &0.387 &0.330 &0.417 &0.491 & &0.290 &0.208 &0.316 &0.458 & &0.830 &0.802 &0.852 &0.868    \\
MuRE   &$\times$ &$\times$ &$\times$ & &0.458 &\underline{0.421} &0.471 &0.525 & &0.313 &0.226 &0.340 &0.489 & &0.940 &0.938 &0.941 &0.943   \\
MuRP   &$\times$ &$\surd$ &$\times$ &  &0.465 &0.420 &\underline{0.484} &0.544 & &0.323 &0.235 &0.353 &0.501 & &\underline{0.941} &\underline{0.939} &0.942 &0.944  \\
AttE   &$\times$ &$\times$ &$\times$ &  &0.456 &0.419 &0.471 &0.526 & &0.311 &0.223 &0.339 &0.488 & &0.939 &0.937 &0.941 &0.942   \\
AttH   &$\times$  &$\surd$ &$\times$ &  &\underline{0.466} &0.419 &\underline{0.484} &\underline{0.551} & &0.324 &0.236 &0.354 &0.501 & &\underline{0.941} &\textbf{0.940} &0.942 &0.945  \\
CompGCN &$\surd$ &$\times$ &$\surd$ & &0.434 &0.406 &0.445 &0.485  & &0.313 &0.227 &0.342 &0.487 & &0.934 &0.930 &0.936 &0.939   \\
\hline
DistMult &$\surd$ &$\times$ &$\times$ &  &0.420 &0.393 &0.429 &0.472 & &0.315 &0.230  &0.344  &0.485 & &0.935 &0.927 &\underline{0.943} &0.945   \\
DistMult-H &$\surd$ &$\surd$ &$\times$ &  &0.433 &0.401 &0.441 &0.497 & &0.322 &0.235  &0.354  &0.496  & &0.936 &0.928 &\underline{0.943} &0.946    \\
DistMult-FFHR &$\surd$ &$\surd$ &$\surd$ &  &0.438 &0.395 &0.457 &0.519 & &0.322   &0.235  &0.352  &0.496 & &0.938 &0.932 &\underline{0.943} &\underline{0.948}    \\
\hline
ComplEx  &$\surd$ &$\times$ &$\times$ &  &0.429 &0.406 &0.436 &0.473  & &0.319 &0.233 &0.350  &0.493 & &0.939 &0.936 &0.942 &0.944   \\
ComplEx-H &$\surd$ &$\surd$ &$\times$  &  &0.442 &0.408 &0.456 &0.507 & &0.325 &0.237 &0.356 &0.500 & &\underline{0.941} &\underline{0.939} &0.942 &0.944  \\
ComplEx-FFHR &$\surd$ &$\surd$ &$\surd$  &  &0.459 &0.420 &0.476 &0.530 & &0.326 &0.236  &0.359  &0.503 & &\textbf{0.943} &\textbf{0.940} &\textbf{0.944} &0.946  \\
\hline 
DualE  &$\surd$ &$\times$ &$\times$ &  &0.385 &0.369 &0.388 &0.413  & &0.290 &0.205 &0.318 &0.462 & &0.935 &0.932 &0.937 &0.939  \\
DualE-H &$\surd$ &$\surd$ &$\times$  &  &0.422 &0.371 &0.451 &0.512 & &0.314 &0.227 &0.345 &0.486 & &0.935 &0.931 &0.938 &0.941   \\
DualE-FFHR &$\surd$ &$\surd$ &$\surd$  &  &0.440 &0.383 &0.474 &0.536 & &0.314 &0.225  &0.348  &0.490 & &0.937 &0.930 &\underline{0.943} &0.945  \\
\hline
RESCAL  &$\surd$ &$\times$ &$\times$  &  &0.449 &0.420 &0.461 &0.503 & &0.337 &0.247 &0.369 &0.514 & &0.927 &0.912 &0.939 &\textbf{0.951}   \\
RESCAL-H  &$\surd$ &$\surd$ &$\times$  &  &0.459 &0.419 &0.480 &0.532 & &\underline{0.344} &\underline{0.255} &\underline{0.378} &\textbf{0.522} & &0.939 &0.936 &0.942 &0.944   \\
RESCAL-FFHR &$\surd$ &$\surd$ &$\surd$  &   &\textbf{0.468} &\textbf{0.422} &\textbf{0.490} &\textbf{0.552} & &\textbf{0.345} &\textbf{0.256}    &\textbf{0.379}    &\underline{0.521} & &0.939 &0.936 &0.942 &0.944  \\
\hline
\end{tabular}
}
\label{table_low}
\end{table*}

\subsubsection{Training and Optimization.}
In this work, we refer to \cite{pmlr-v80-lacroix18a} and use the fully multiclass log-loss as the objective function:
\begin{equation*}
    L((h,r,t)) = -s(h,r,t) + log\left[\sum_{t'\in \mathcal{E}} exp(s(h,r,t'))\right].
\end{equation*}
Also, We refer to previous work \cite{chami2020low}\cite{10.1145/3442381.3450118} for hyperbolic model optimization.

\section{Experiments and Analyses}
In this section, we conduct experiments to evaluate the performance of our proposals. Our experiments are intended to address the following major questions: 
\begin{itemize}[leftmargin=*]
    \item \textbf{(RQ1)} How does FFHRs perform comparing with existing methods, \textit{w.r.t.} Euclidean-based methods and hyperbolic distance-based methods?
    \item \textbf{(RQ2)} Whether hyperbolic space and HIN contribute to capturing hierarchical data?
    \item \textbf{(RQ3)} Is HIN superior over hyperbolic distance in capturing complex relation patterns?
    \item \textbf{(RQ4)} Does the proposed component FPM-GCN indeed reduce approximation error and boost performance?
    \item \textbf{(RQ5)} How do different hyper-parameter settings (e.g. dimension, regularization coefficients) affect performance?
\end{itemize}

\subsection{Experimental Settings}
\subsubsection{Datasets.} We use three standard benchmarks WN18 \cite{10.5555/2999792.2999923}, WN18RR \cite{dettmers2018conve} and FB15k-237 \cite{toutanova2015observed} for fair comparison. WN18 is a subset of WordNet, which contains rich hierarchical relations, such as "hypernym", "part$\_$of", and many inverse relations. WN18RR is a subset of WN18, where the inverse relations are deleted. FB15k-237 is a subset of Freebase, containing general world knowledge. Most relations of it are non-hierarchical, e.g., "born-in or nationality", while still having some hierarchical relations, e.g., "part-of").
 The statistics of the datasets are presented in Table \ref{table_dataset}.
\subsubsection{Baselines. }
In the following, we compare our method to state-of-the-art methods of different categories: \\
\textbf{Distance-based Models}:
\begin{itemize}[leftmargin=*]
    \item \textbf{RotatE} \cite{sun2018rotate} is a classical Euclidean distance-based method, which formulates relations of various patterns as 2D rotations from head entities to tail entities.
    \item \textbf{Rotate3D} \cite{10.1145/3340531.3411889} is an extensive version of RotatE, which generalizes 2D rotations to 3D rotations and benefits multi-hop reasoning.
\end{itemize}
\textbf{Inner-Product-based Models}:
\begin{itemize}[leftmargin=*]
    \item \textbf{ConvE} \cite{dettmers2018conve} is a representative inner-product-based method, which leverages 2D convolution and fully-connected neural network to model the complex interaction between entities and relations.
    \item \textbf{TuckER} \cite{balazevic2019tucker} is a straightforward but powerful linear model based on Tucker decomposition of the binary tensor representation of knowledge graph triples.
    \item \textbf{QuatE} \cite{NEURIPS2019_d961e9f2} utilizes hypercomplex-valued embeddings with three imaginary components to represent entities and relations, which is an extensive version of ComplEx.
    \item \textbf{LowFER} \cite{pmlr-v119-amin20a} proposes a factorized bilinear pooling model for better fusion of entities and relations, which naturally generalizes TuckER model.
    \item \textbf{AutoSF} \cite{DBLP:conf/icde/ZhangYDC20} utilizes technology of automated machine learning to set up a search space, where they automatically search an embedding model for a given KG.
\end{itemize}
\textbf{Distance-based hyperbolic models and variants}:
\begin{itemize}[leftmargin=*]
    \item \textbf{MuRP}, \textbf{MuRE} \cite{balazevic2019multi} is the first one to utilize hyperbolic embedding to model the hierarchical data in KG. The score function is based on hyperbolic distance function.
    \item \textbf{AttH}, \textbf{AttE} \cite{chami2020low} represents relations as parameterized operations (rotations and reflections) on hyperbolic space, which better models the complex patterns of relations.
    \item \textbf{ConE} \cite{bai2021cone} represents entities and queries as Cartesian products of two-dimensional cones, which shows great power in multi-hop reasoning over KGs.
\end{itemize}
\textbf{GNN-based models}:
\begin{itemize}[leftmargin=*]
    \item \textbf{CompGCN} \cite{vashishth2020compositionbased} is a general framework which incorporates GCN with knowledge graph embedding by leveraging a variety of entity-relation composition operations.
    \item \textbf{M$^2$GNN} \cite{10.1145/3442381.3450118} embeds multi-relational KGs in a mixed-curvature space for knowledge graph completion.
\end{itemize}
\subsubsection{Evaluation Metrics.}
We follow a standard evaluation setting that filters out existing triples during testing \cite{10.5555/2999792.2999923} and use two conventional metrics across the link prediction literature:
\begin{itemize}
\item MRR: Mean reciprocal rank of correct entities.
\item H@K, K$\in$\{1,3,10\}: measures the correct ratio that a true triple is ranked
within the top k candidate triples.
\end{itemize}

\subsubsection{Implementation Details.}
We entirely follow previous work \cite{chami2020low} in dataset division and data augmentation. In low dimension case, for a fair comparison, the embedding dimension of all methods including baselines are equally set to 32. As our experimental settings are entirely the same as \cite{chami2020low}, some baseline results are directly taken from \cite{chami2020low}. We also reproduce a part of important ones and get similar results with [9]. In high dimension case, we set the dimension for our methods and their variants to 512. For other baselines, we do not constrain their dimension, but report their best performance in arbitrary dimension in the range [200, 2000] referring to their original papers.\par
Regarding to hyper-parameters, for our FFHRs and their variants, we utilize grid search to find the best with batch size $\in \{100,200,500,1000,2000\}$, learning rate $\in \{0.005, 0.01, 0.05, 0.1, 0.5\}$, regularization coefficients $\in \{0.005, 0.01, 0.05, 0.1, 0.5\}$, number of attention heads $\in \{1, 2, 4, 8\}$, number of GCN layer $\in \{1, 2\}$. The hyper-parameters of baselines are referring to original papers or validated in our experiments. \par
For all our methods and their Euclidean counterparts, We adopt Adagrad as optimizer and initialize embeddings with default Xavier uniform initializer. Note that another advantage of inner-product-based methods is that they can use various regularizers to boost performance, while distance-based methods can not. In this paper, we employ DURA \cite{conf/nips/ZhangC020} as regularizer. For a fair comparison, we reimplement DualE, DualE-H, DualE-FFHR without type constraint used in the original paper\cite{Cao_Xu_Yang_Cao_Huang_2021}. We implement the proposed methods in PyTorch and conduct all the experiments on NVIDIA Tesla V100 GPU.

\begin{table*}[t]
\centering

\renewcommand{\arraystretch}{1.3}
\caption{KGC performance comparison in high dimension case (d=512), where ``IP'', "HS", "GCN" denote whether using Inner product, Hyperbolic Space, GCN respectively. The best results are in bold and the second best are underlined.}
\resizebox{\textwidth}{!}{
\begin{tabular}{l c c c c c c c c c c c c c c c c c c }
\hline
  &\multicolumn{3}{c}{characteristic} &   &\multicolumn{4}{c}{WN18RR} & &\multicolumn{4}{c}{FB15k-237}  & &\multicolumn{4}{c}{WN18}\\
\cline{2-4} \cline{6-9} \cline{11-14}\cline{16-19}
model &IP? &H? &GCN? & &MRR &H@1 &H@3 &H@10 &  &MRR &H@1 &H@3 &H@10 & &MRR &H@1 &H@3 &H@10 \\
\hline
ConvE  &$\surd$ &$\times$ &$\times$  & &0.430 &0.400 &0.440 &0.520 & &0.325 &0.237 &0.356 &0.501 & &0.943 &0.935 &0.946 &0.956   \\
TuckER &$\surd$ &$\times$ &$\times$ &  &0.470 &0.443 &0.482 &0.526 & &0.358 &0.266 &0.394 &0.544 & &\textbf{0.953} &\textbf{0.949} &\underline{0.955} &0.958   \\
QuatE &$\surd$ &$\times$ &$\times$ & &0.488 &0.438 &\underline{0.508} &0.582 & &0.348 &0.248 &0.382 &0.550 & &0.950 &0.945 &0.954 &0.959   \\
AutoSF &$\surd$ &$\times$ &$\times$ & &0.490 &0.451 &- &0.567 & &0.360 &0.267 &- &0.552 & &\underline{0.952} &\underline{0.947} &- &\underline{0.961} \\
RotatE &$\times$ &$\times$ &$\times$ & &0.476 &0.428 &0.492 &0.571 & &0.338 &0.241 &0.375 &0.533 & &0.947 &0.938 &0.953 &\underline{0.961}   \\
Rotate3D  &$\times$ &$\times$ &$\times$ & &0.489 &0.442 &0.505 &0.579 & &0.347 &0.250 &0.385 &0.543 & &0.951 &0.945 &0.953 &\underline{0.961}  \\
LowFER &$\surd$ &$\times$ &$\times$ & &0.465 &0.434 &0.479 &0.526 & &0.359 &0.266 &0.396 &0.544 & &0.950 &0.946 &0.952 &0.958   \\
MuRP   &$\times$ &$\surd$ &$\times$ & &0.481 &0.440 &0.495 &0.566 & &0.335 &0.243 &0.367 &0.518 & &0.947 &0.942 &0.951 &0.954  \\
AttH   &$\times$ &$\surd$ &$\times$ & &0.486 &0.443 &0.499 &\underline{0.573} & &0.348 &0.252 &0.384 &0.540 & &0.948 &0.944 &0.950 &0.953 \\
ConE  &$\times$ &$\surd$ &$\times$ & &0.496 &0.453 &0.515 &0.579 & &0.345 &0.247 &0.381 &0.540 & &- &- &- &-\\
CompGCN &$\times$ &$\times$ &$\surd$ & &0.479 &0.443 &0.494 &0.546 & &0.355 &0.264 &0.390 &0.535 & &0.945 &0.941 &0.948 &0.951  \\
M$^2$GNN &$\times$ &$\surd$ &$\surd$ & &0.485 &0.444 &0.498 &0.572 & &\underline{0.362} &\underline{0.275}  &0.398  &\textbf{0.565}  & &- &- &- &-   \\
\hline
DistMult-FFHR &$\surd$ &$\surd$ &$\surd$ & &0.459  &0.418 &0.473 &0.544  & &0.360 &0.266 &0.395 &0.547 & &0.947 &0.942 &0.949 &0.953   \\
ComplEx-FFHR &$\surd$ &$\surd$ &$\surd$ & &0.488 &\underline{0.446} &0.504 &0.568 & &\underline{0.362} &0.268 &\underline{0.399} &0.551 & &\textbf{0.953} &\underline{0.947} &\textbf{0.956} &\textbf{0.963}   \\
DualE-FFHR &$\surd$ &$\surd$ &$\surd$  &  &0.486 &0.446 &0.504 &0.564 & &0.353 &0.259  &0.390  &0.541 & &0.951 &0.946 &\underline{0.955} &\underline{0.961}  \\
RESCAL-FFHR &$\surd$ &$\surd$ &$\surd$ & &\textbf{0.500} &\textbf{0.461} &\textbf{0.515} &\textbf{0.577} & &\textbf{0.370} &\textbf{0.277} &\textbf{0.407}  &\underline{0.552}   & &0.951 &0.945 &\underline{0.955} &\underline{0.961}    \\
\hline
\end{tabular}
}
\label{result_high}
\end{table*}


\begin{table*}[!t]
\centering
\renewcommand{\arraystretch}{1.3}
\caption{Hits@10 on WN18RR with d=32. Higher Khs$_G$ and lower $\xi_G$ means more hierarchical structure.}

\begin{tabular}{l c c c c c} 
\hline
Relation &Khs$_G$  &$\xi_G$ &RESCAL &RESCAL-H  & Rel Impro\\
\hline
member\_meronym &1.00  &-2.90 &0.330 &0.368  &+11.5\% \\
hypernym &1.00 &-2.46 &0.175 &\textbf{0.211} &\textbf{+20.6\%} \\
has\_part &1.00 & -1.43 &0.279 &0.308 &+10.4\% \\
instance\_hypernym &1.00 & -0.82 &0.516 &0.570 &+10.5\% \\
member\_of\_domain\_region &1.00 &-0.78 &0.288 &\textbf{0.346} &\textbf{+20.1\%} \\
member\_of\_domain\_usage &1.00 &-0.74 &0.375 &0.438 &+16.8\% \\
synset\_domain\_topic\_of &0.99 &-0.69 &0.395 &0.439 &+11.1\% \\
\hline
also\_see &0.36 &-2.09 &0.625 &0.616 &-1.4\% \\
derivationally\_related\_form &0.07 &-3.84 &0.959 &0.963 &+0.4\% \\
similar\_to  &0.07 &-1.00 &1.000 &1.000 &+0.0\% \\
verb\_group &0.07 &-1.00 &0.974 &0.974 &+0.0\% \\
\hline
\end{tabular}

\label{table_relation}

\end{table*}

\begin{table*}[!t]
\centering
\renewcommand{\arraystretch}{1.3}
\caption{Results on different relation categories (categorized by \cite{10.5555/2893873.2894046}), in terms of dataset FB15k-237 and d=32.}
\begin{tabular}{l c c c c c c c c c c c c }
\hline
  &    &\multicolumn{2}{c}{MuRP} & &\multicolumn{2}{c}{AttH}  & &\multicolumn{2}{c}{RESCAL-H-dist}& &\multicolumn{2}{c}{RESCAL-H}\\
\cline{3-4} \cline{6-7} \cline{9-10} \cline{12-13}
 & &MRR &H@10 & &MRR &H@10 & &MRR &H@10 & &MRR &H@10  \\
\hline
\multirow{4}{*}{Head Prediction}
& 1-1 &\textbf{0.477} &0.568 &  &0.467 &\textbf{0.573} &   &0.458 &0.531 &   &0.437 &\textbf{0.573}\\
& 1-N &0.088 &0.170 &  &0.073 &0.150 &   &0.094 &0.178 &   &\textbf{0.103} &\textbf{0.193}\\
& N-1 &0.445 &0.643 &  &0.439 &0.630 &   &0.447 &0.630 &   &\textbf{0.449} &\textbf{0.645}\\
& N-N &0.246 &0.449 &  &0.243 &0.439 &   &0.255 &0.456 &   &\textbf{0.272} &\textbf{0.471}\\
\hline
\multirow{4}{*}{Tail Prediction}
& 1-1 &\textbf{0.474} &0.557 &   &0.453 &\textbf{0.573} &   &0.470 &0.552 &   &0.439 &0.557\\
& 1-N &0.733 &0.858 &   &0.730 &0.852 &   &0.733 &0.851 &   &\textbf{0.760} &\textbf{0.865}\\
& N-1 &0.059 &0.131 &   &0.060 &0.111 &   &0.070 &0.136  & &\textbf{0.079} &\textbf{0.156}\\
& N-N &0.352 &0.581 &   &0.350 &0.577 &   &0.361 &0.589 &   &\textbf{0.382} &\textbf{0.601}\\
\hline
\end{tabular}
\label{table_NN}
\end{table*}

\subsection{Performance Comparisons (RQ1)}
\subsubsection{Results in Low Dimension.} Table \ref{table_low} presents the results with the low dimension (d=32). We make the following observations: (1) Our methods consistently achieve the best performance on all benchmarks, demonstrating the effectiveness of the proposed HIN and FPM-GCN. (2) Comparing with Euclidean counterparts, our hyperbolic models without FPM-GCN (*-H) improve by an average of 2.8\%, 2.0\%, and 0.5\% points in terms of MRR on WN18RR, FB15k-237, and WN18 respectively. This is consistent with our expectations that hyperbolic space is more suitable for modeling hierarchical data, especially in low dimension case. Notice that WN18RR contains more hierarchical data structures, where our hyperbolic models achieve more improvements over their Euclidean counterparts on it. (3) Our methods also significantly outperform previous distance-based hyperbolic methods (\textit{e.g.} AttH, MuRP) on FB15k-237, as our flexible inner product function is beneficial to capture complex relations. (4) With few exceptions, leveraging FPM-GCN would improve model performance. \par
\subsubsection{Results in High Dimension.}
Performance of our hyperbolic models and their variants with $d=512$ are presented in table \ref{result_high}. For other baselines, we do not constrain their dimension but report their best performance in arbitrary dimension in
the range [200, 2000] referring to their original papers. The missing baselines are due to the missing open source codes. With few exceptions, we can find that our methods still achieve state-of-the-art in all benchmarks and improve a lot in handling complex data patterns (\textit{e.g.} FB15K-237), while other methods fall short in such cases. It may be caused by their inflexible score function and approximation error.

\subsection{Case Study (RQ2, RQ3)}
\subsubsection{Case Study on capturing hierarchical data (RQ2)}
For a deeper understanding of the superiority of hyperbolic space in modeling hierarchical data, we measure performance in terms of different relation types on WN18RR. Following \cite{chami2020low}, we use Krackhardt hierarchy score (Kh$s_G$) and estimated curvature ($\xi_G$) for each relation. Lower $\xi_G$ and higher Kh$s_G$ measure the extent of global and local tree structure respectively. We compute the averaged Hit@10 over 10 runs for each relation in low dimension case (d=32). As shown in Table \ref{table_relation}, tree-liked relations (\textit{e.g.} "hypernym", "has\_part") receive greater improvement with FFHR. Inversely, for non-hierarchical relations (\textit{e.g.} "also\_see", "similar\_to"), showed in lower part of Table \ref{table_relation}, hyperbolic model achieves only slightly or even negative improvements. \par

\subsubsection{Case Study on capturing complex relation patterns (RQ3)}
Following \cite{10.5555/2893873.2894046}, we divide the relations on FB15k-237 into four categories: one-to-one (1-1), one-to-many (1-N), many-to-one (N-1), many-to-many (N-N). We compare our model RESCAL-H with hyperbolic-distance-based models, AttH, MuRP, and RESCAL-H-dist (the variation of RESCAL-H that only replace HIN with hyperbolic distance function \cite{balazevic2019multi}). Table \ref{table_NN} shows that in contrast to hyperbolic distance-based methods (AttH, MuRP, and RESCAL-H-dist), RESCAL-H achieves impressive improvement on complex relations (1-N, N-1, N-N). For simple relations (1-1), we can find hyperbolic distance function performs better. This result clearly validates the superiority of hyperbolic inner product in capturing complex data patterns. In the real world and our benchmarks, complex relations (1-N, N-1, N-N) are the majority, which demonstrates the superiority to make use of hyperbolic inner product function.

\begin{figure}
	\centering

     \includegraphics[width=0.8\columnwidth]{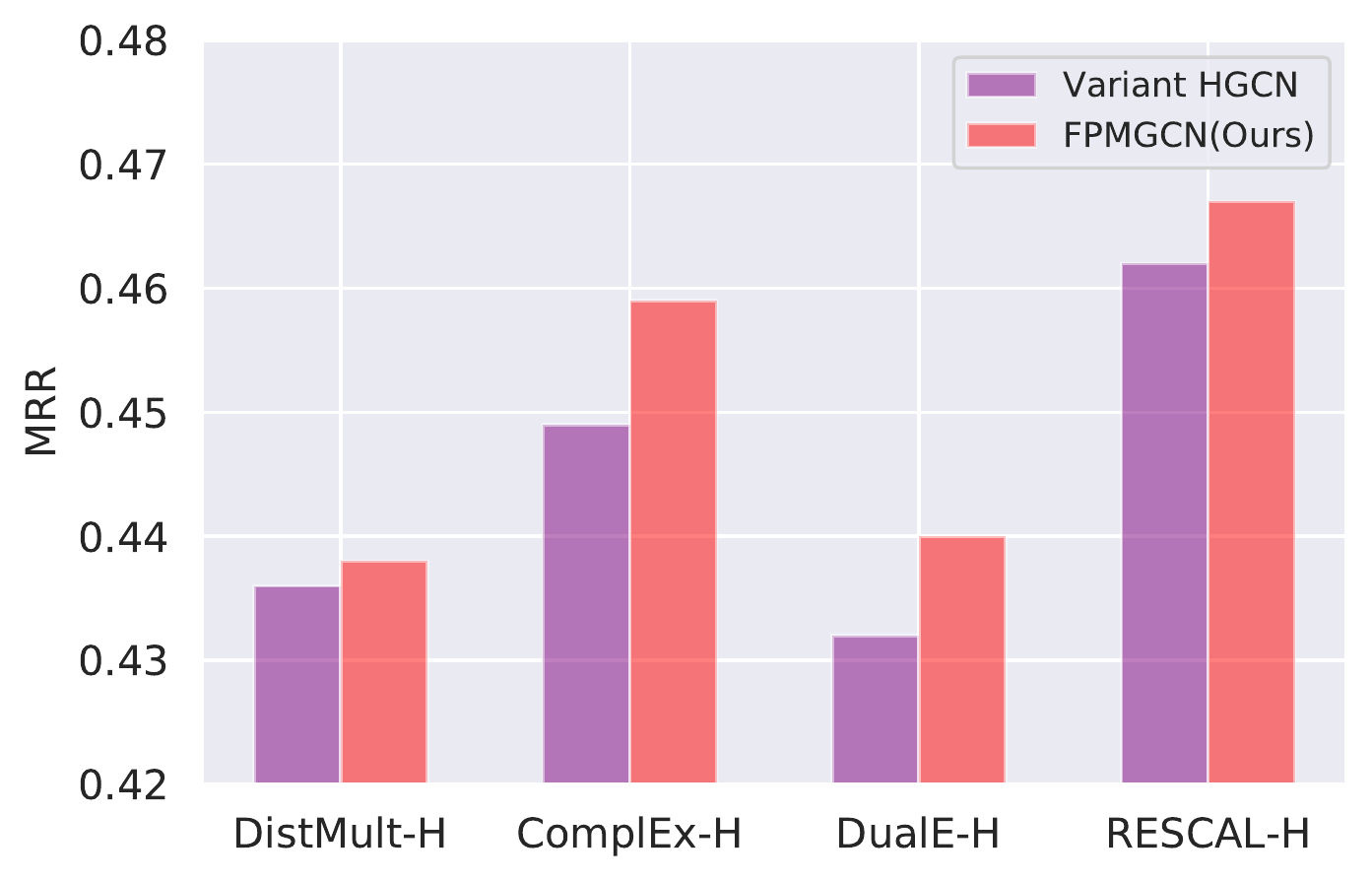}
    	
	\caption{Comparison between FPM-GCN and variants HGCN on WN18RR with d=32.}
	\label{figure_ablation}
\end{figure}

\begin{figure}
	\centering
     \includegraphics[width=0.8\columnwidth]{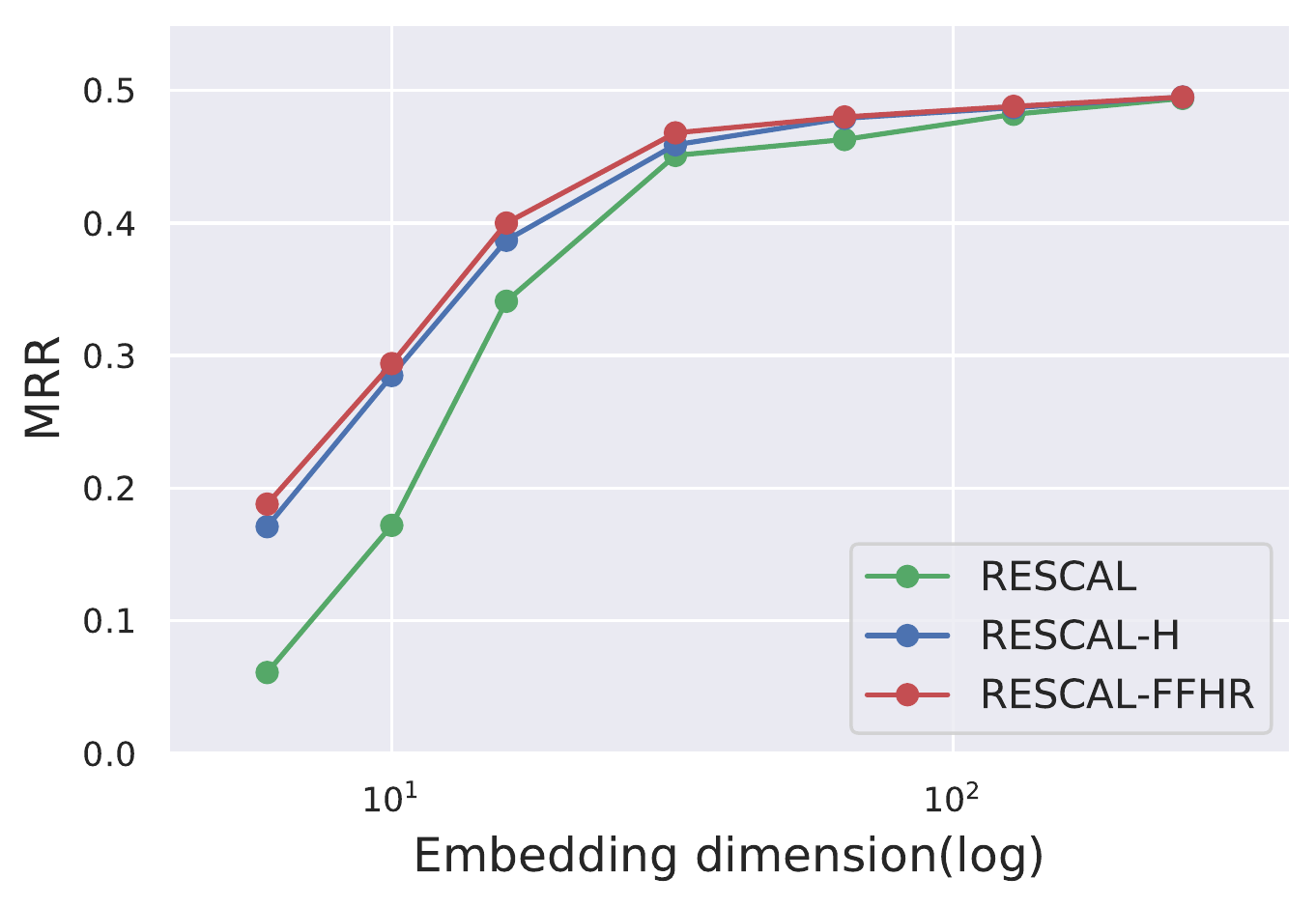}
	\caption{MRR as a function of the embedding dimension with d $\in \{6,10,16,32,64,128,256\}$.}
	\label{figure_dimension}
\end{figure}

\subsection{Ablation Study (RQ4, RQ5)}
\subsubsection{Ablation Study on FPM-GCN (RQ4)}
To give a fair comparison between our proposed FPM-GCN and previous hyperbolic GCN which use tangent space approximation, in this section, we implement four variations DistMult-V-HGCN, ComplEx-V-HGCN, DualE-V-HGCN, and RESCAL-V-HGCN, where only feature transformation module and aggregation module are performed via tangent space approximation, while all other components are the same as FPM-GCN. Concretely, the tangent space based modules are defined as follow:\par
\textbf{Feature transformation module of HGCN}: 
\begin{equation*}
    \mathbf{m}_i^{l,\mathcal{B}_c} = \mathbf{W}_r^l \otimes_c \mathbf{x}_i^{l-1, \mathcal{B}_c}
\end{equation*}
where $\mathbf{W}_r^l$ are general matrices.\\
\textbf{Neighborhood aggregation module of HGCN}:
\begin{equation*}
    \mathbf{x}_i^{k,l,\mathcal{B}_c}=exp_0\left( \sum_{j\in N(i)}\frac{v_{i,j}^k\cdot log_0(\mathbf{m}_j^{l,\mathcal{B}_c})}{\sum_{j\in N(i)} v_{i,j}^k } \right)
\end{equation*}
where $v_{i,j}^k$ is k-th attention weight and $N(i)$ is the set of one-hop neighborhood of entity $i$. \par
Since these operations are defined in the tangent space of the origin, they don’t have clear meanings in hyperbolic space. And the hyperbolic structure between entities will also be destroyed using these operations.\par 
Here we choose HGCN for comparison rather than H2HGNN\cite{Dai_2021_CVPR}, as the latter is designed for single-relation graph on Lorentz model, which can't adapt to our score function defined on Poincar\'{e} ball model. \par 
Fig. \ref{figure_ablation} presents the MRR results on WN18RR with d=32. Our FPM-GCN consistently outperforms variant HGCN for all base models by an average of 1.5\%, which clearly validates the superiority of our FPM-GCN. The key reason is that the operations we defined are entirely based on hyperbolic space without approximation. 

\subsubsection{Ablation Study on Dimension (RQ5)}
To investigate the role of the embedding dimension, we report the MRR on WN18RR of our models (RESCAL, RESCAL-H, RESCAL-FFHR) with different dimension as shown in Fig. \ref{figure_dimension}. The results show that RESCAL-H and RESCAL-FFHR consistently outperform their Euclidean counterpart RESCAL and improve significantly in low dimension case. That confirms our expectation that hyperbolic embeddings are more expressive in low dimension case due to the geometric characteristics. Besides that, the FPM-GCN shows its improvement as an encoder across a broad range of dimensions compared to RESCAL-H.


\section{Conclusion}
In this paper, with rigorous mathematical reasoning, we propose two important operations in hyperbolic space: (1) FPM-GCN, which does not require tangent space approximation; (2) HIN, a hyperbolic generalization of Euclidean inner product which is more flexible than hyperbolic distance in modeling complex data patterns. We further present a fully and flexible hyperbolic representation embedding framework FFHR, which is able to transfer recent Euclidean-based advances to hyperbolic space. We demonstrate it by instantiating FFHR with four representation KGC methods and conduct extensive experiments to validate the effectiveness of the proposals.  \par
For further work, rigorous proof of the superiority of inner product and more discussion between the two score functions are promising. Meanwhile, mixed spaces have a more powerful representation ability and attract considerable attention recently. It would be valuable to generalize our GCN module and inner product function to mix space.

\ifCLASSOPTIONcaptionsoff
  \newpage
\fi



%
\bibliographystyle{IEEEtran}
\bibliography{mybibfile}

%

\begin{IEEEbiography}[{\includegraphics[width=1in,height=1.25in,clip,keepaspectratio]{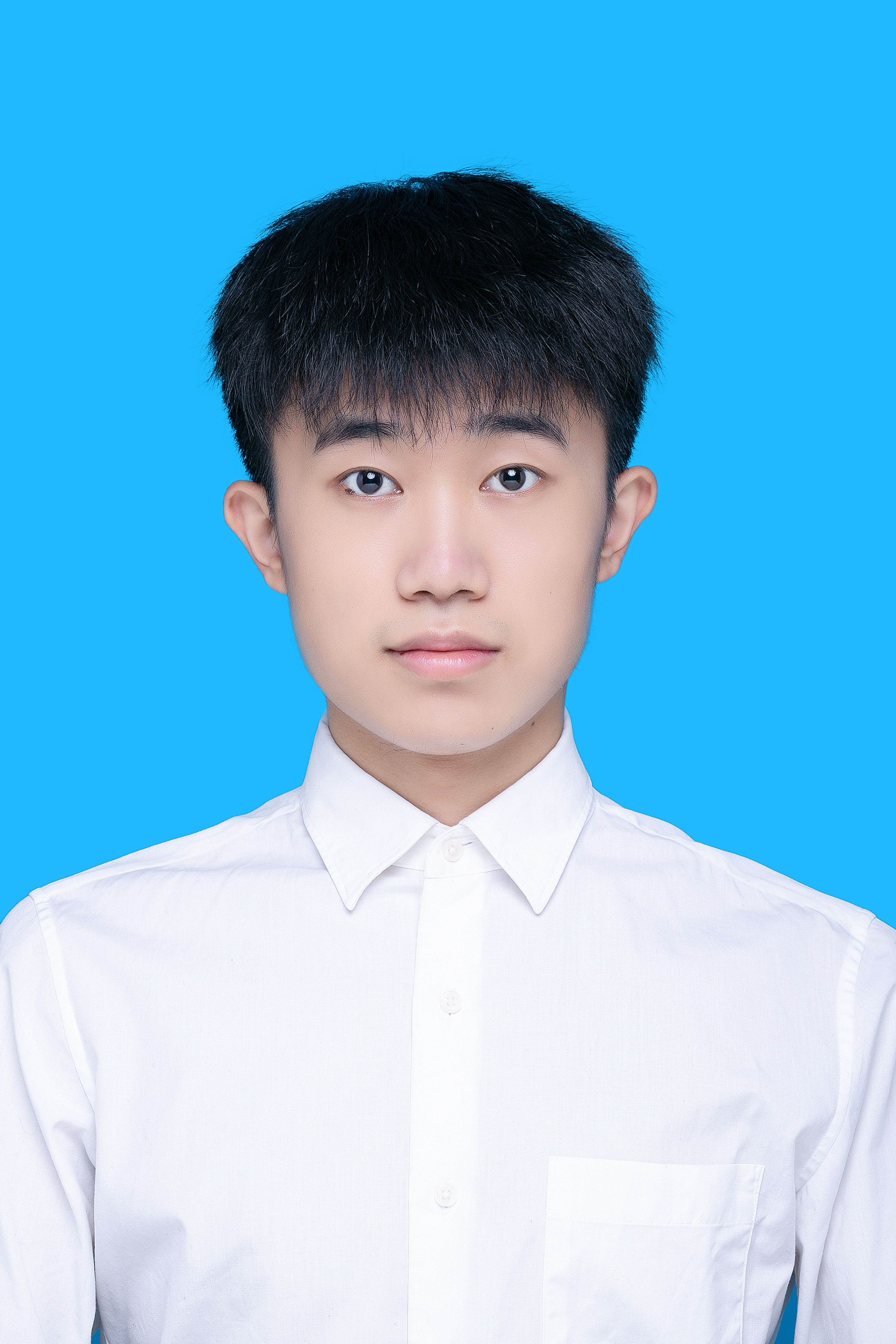}}]{Wentao Shi}
is currently working towards the
master degree at University of Science and Technology of China (USTC), Hefei, China. His research interests include data mining and recommender systems.
\end{IEEEbiography}

\begin{IEEEbiography}[{\includegraphics[width=1in,height=1.25in,clip,keepaspectratio]{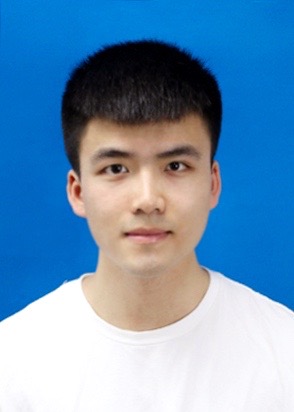}}]{Junkang Wu}
is currently working towards the
master degree at University of Science and Technology of China (USTC), Hefei, China. His research interests include data mining and recommender systems.
\end{IEEEbiography}

\begin{IEEEbiography}[{\includegraphics[width=1in,height=1.25in,clip,keepaspectratio]{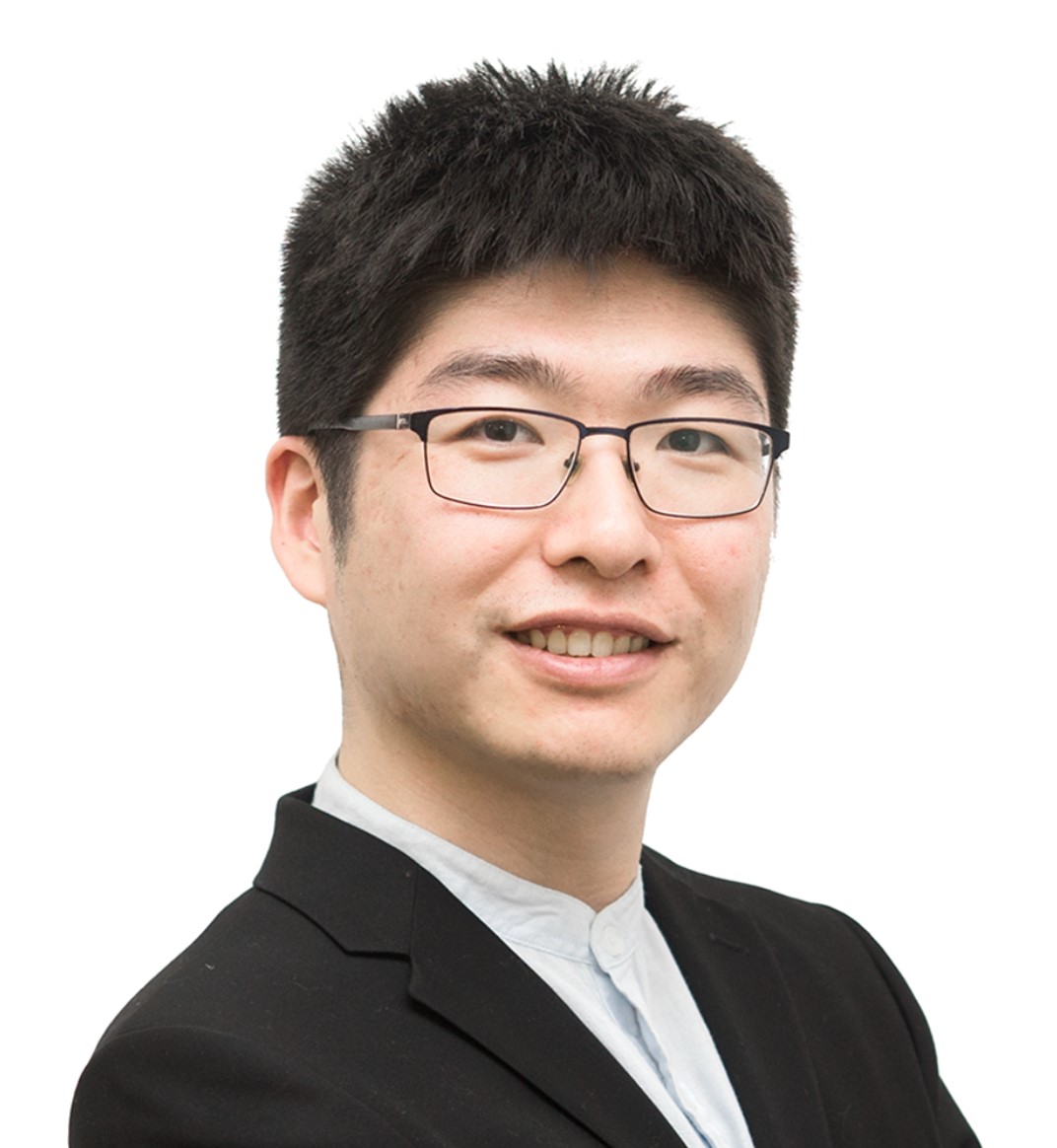}}]{Xuezhi Cao} is a senior researcher in Meituan’s
NLP team. He obtained his Ph.D. degree from Shanghai Jiao Tong University in 2018. His research interests include knowledge graph, recommender system, and social network. He has over 15 publications in top conferences including SIGIR, WWW, AAAI, etc.
\end{IEEEbiography}

\begin{IEEEbiography}[{\includegraphics[width=1in,height=1.25in,clip,keepaspectratio]{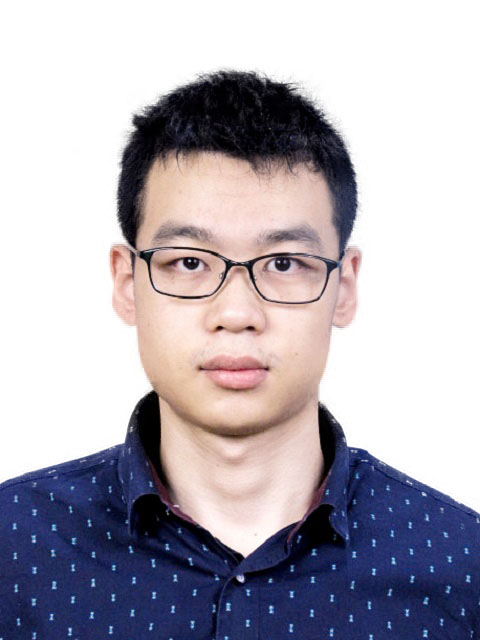}}]{Jiawei Chen}is a professor at Zhejiang University. He received Ph.D. in Computer Science from Zhejiang University in 2020. His research interests include information retrieval, data mining, and causal inference. He has published over ten academic papers on international conferences and journals such as WWW, AAAI, SIGIR, CIKM, ICDE and TOIS. Moreover, he has served as the PC/SPC member for top-tier conferences including SIGIR, WWW, WSDM, ACMMM, AAAI, IJCAI and the invited reviewer for prestigious journals such as TNNLS, TKDE, TOIS.
\end{IEEEbiography}

\begin{IEEEbiography}[{\includegraphics[width=1in,height=1.25in,clip,keepaspectratio]{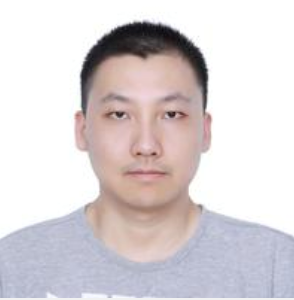}}]{Wei Wu} is now a technical leader in Meituan. He is leading a team focus on AI technologies such as knowledge graph, NLP, and information retrieval. Before this, Dr. Zhang was a researcher in Microsoft Research Asia. He obtained his Ph.D. degree in computer science, and is supervised jointly by University of Science and Technology of China and Microsoft Research Asia. He has published over 50 top-tier international conference papers and journal articles including KDD, WWW, AAAI, and IJCAI. He has received the best paper award in ICDM2013 and CIKM2020. He has long served as the reviewers on top-tier international conferences and journals, such as KDD, WWW, and TKDE.
\end{IEEEbiography}

\begin{IEEEbiography}[{\includegraphics[width=1in,height=1.25in,clip,keepaspectratio]{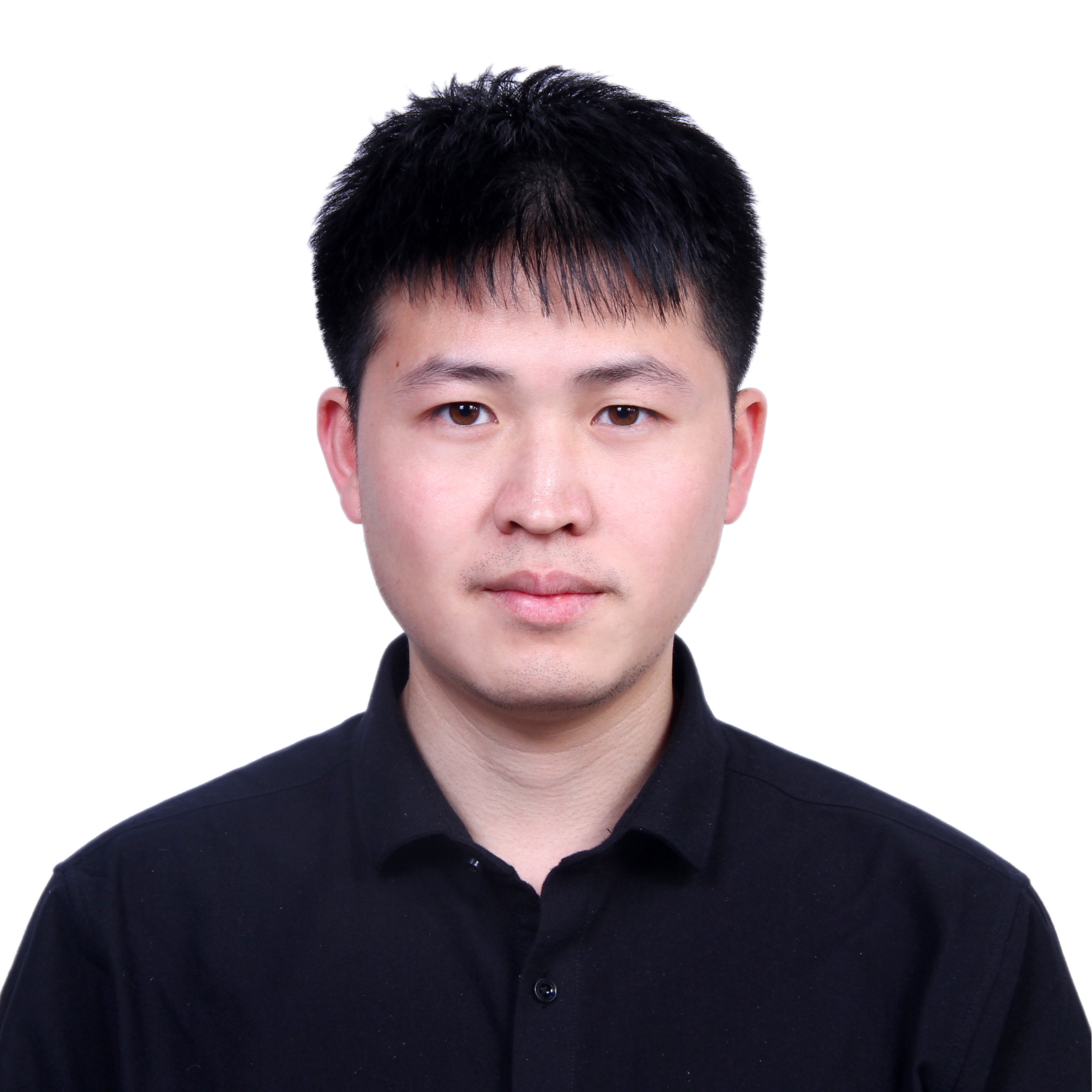}}]{Weiqiang Lei} is a professor in Sichuan University. He received his Ph.D. degree from National University of Singapore in 2019. His research interests focus on conversational AI, inclusive of conversational recommendation, dialogue and QA system, user feedback modeling. He has published relevant papers at top venues such as KDD, WSDM, TOIS, ACL, EMNLP and the winner of ACM MM 2020 best paper award. He has also actively give tutorials on the topic of conversational recommendation at multiple conferences: RecSys 2021, SIGIR 2020, CCL 2020, CCIR 2020
\end{IEEEbiography}

\begin{IEEEbiography}[{\includegraphics[width=1in,height=1.25in,clip,keepaspectratio]{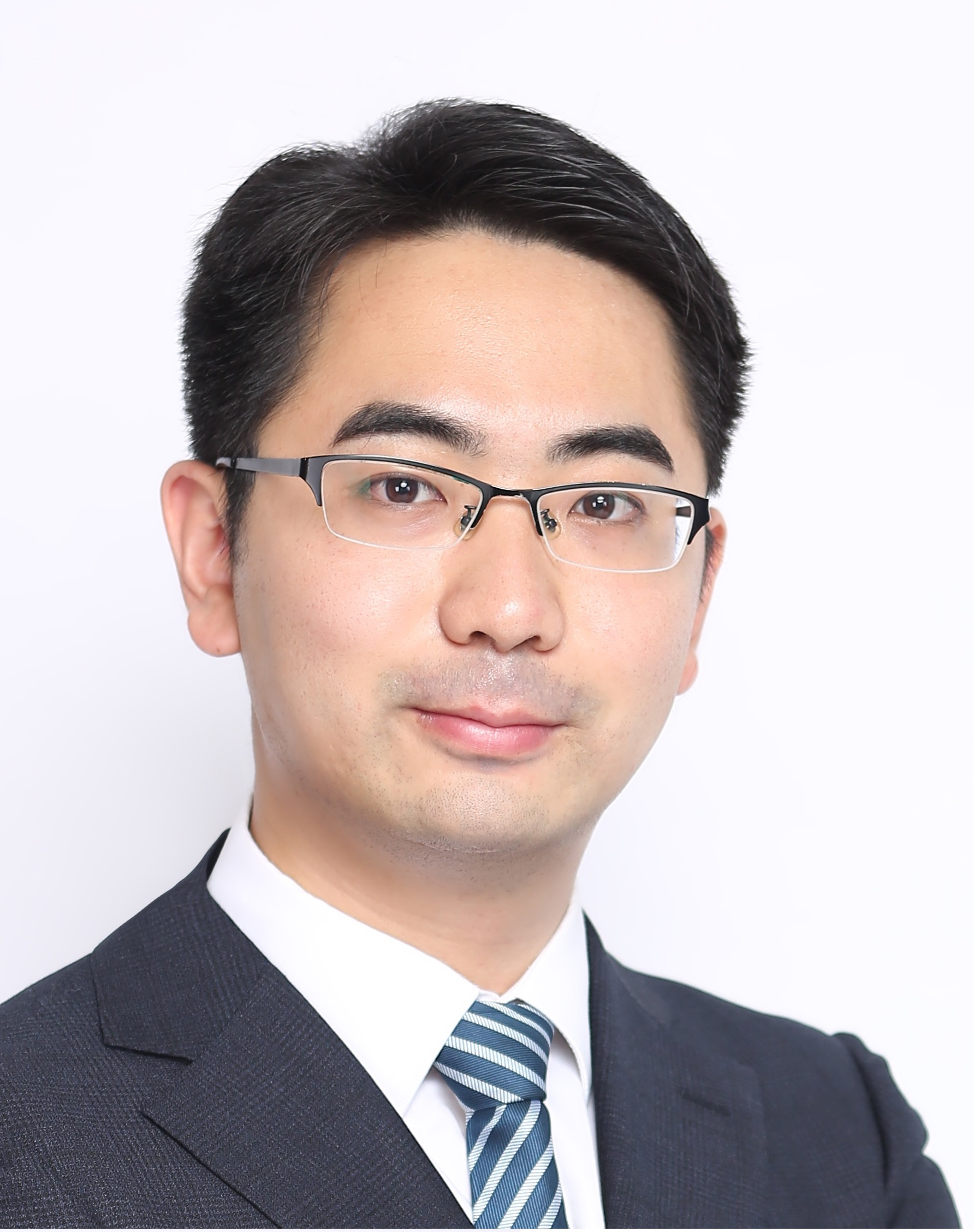}}]{Xiangnan He}
is a professor at the University of Science and Technology of China (USTC). He received his Ph.D. in Computer Science from the National University of Singapore (NUS). His research interests span information retrieval, data mining, and multi-media analytics. He has over 100 publications that appeared in top conferences such as SIGIR, WWW, and KDD, and journals including TKDE, TOIS, and TNNLS. His work has received the Best Paper Award Honorable Mention in SIGIR (2021, 2016) and WWW (2018). He is in the editorial board for several journals including ACM Transactions on Information Systems (TOIS), IEEE Transactions on Big Data (TBD), AI Open, etc. Moreover, he has served as the PC chair of IEEE CCIS 2019 and (senior) PC member for several top conferences including SIGIR, WWW, KDD, MM, WSDM, ICML, etc.
\end{IEEEbiography}





\end{document}